\documentclass[10pt,twocolumn,letterpaper]{article}

\usepackage{iccv}
\usepackage[accsupp]{axessibility}
\usepackage{times}
\usepackage{epsfig}
\usepackage{graphicx}
\usepackage{amsmath}
\usepackage{amssymb}
\usepackage{algorithm}
\usepackage{algorithmic}
\usepackage{multicol}

\usepackage{bm}
\usepackage{booktabs}       
\usepackage{arydshln}    
\usepackage{multirow}
\usepackage{pdfpages}

\usepackage[flushleft]{threeparttable} 
\usepackage{pifont}

\usepackage[pagebackref=true,breaklinks=true,letterpaper=true,colorlinks,bookmarks=false]{hyperref}

\DeclareMathOperator*{\argmin}{arg\,min}
\DeclareMathOperator*{\E}{\mathbb{E}}

\newcommand{\cmark}{\ding{51}}

\def\etal{\emph{et al.}} 

\makeatletter
\renewcommand*{\@fnsymbol}[1]{\ensuremath{\ifcase#1\or 3 \or \dagger\or \ddagger\or
   \mathsection\or \mathparagraph\or \|\or **\or \dagger\dagger
   \or \ddagger\ddagger \else\@ctrerr\fi}}
\makeatother

\iccvfinalcopy 

\def\iccvPaperID{10177} 

\ificcvfinal\fi

\begin{document}
\title{Meta-Learning with Task-Adaptive Loss Function for Few-Shot Learning}

\author{Sungyong Baik\textsuperscript{1} ~ Janghoon Choi\textsuperscript{1,}\thanks{Now at Kookmin University} ~ Heewon Kim\textsuperscript{1} ~ Dohee Cho\textsuperscript{1} ~ Jaesik Min\textsuperscript{2} ~ Kyoung Mu Lee\textsuperscript{1}\vspace{0.2cm}\\
\textsuperscript{1}ASRI, Department of ECE, Seoul National University ~~~~
\textsuperscript{2}Hyundai Motor Group\\
\textsuperscript{1}{\tt\small \{dsybaik, ultio791, ghimhw, jdh12245, kyoungmu\}@snu.ac.kr} ~~~~ \textsuperscript{2}{\tt\small jaesik.min@hyundai.com}
}

\maketitle

\ificcvfinal\thispagestyle{empty}\fi

\begin{abstract}
In few-shot learning scenarios, the challenge is to generalize and perform well on new unseen examples when only very few labeled examples are available for each task.
Model-agnostic meta-learning (MAML) has gained the popularity as one of the representative few-shot learning methods for its flexibility and applicability to diverse problems.
However, MAML and its variants often resort to a simple loss function without any auxiliary loss function or regularization terms that can help achieve better generalization.
The problem lies in that each application and task may require different auxiliary loss function, especially when tasks are diverse and distinct.
Instead of attempting to hand-design an auxiliary loss function for each application and task, we introduce a new meta-learning framework with a loss function that adapts to each task.
Our proposed framework, named Meta-Learning with Task-Adaptive Loss Function (MeTAL), demonstrates the effectiveness and the flexibility across various domains, such as few-shot classification and few-shot regression.
\end{abstract}

\begin{figure}[t]
	\centering
	\includegraphics[width=0.9\linewidth]{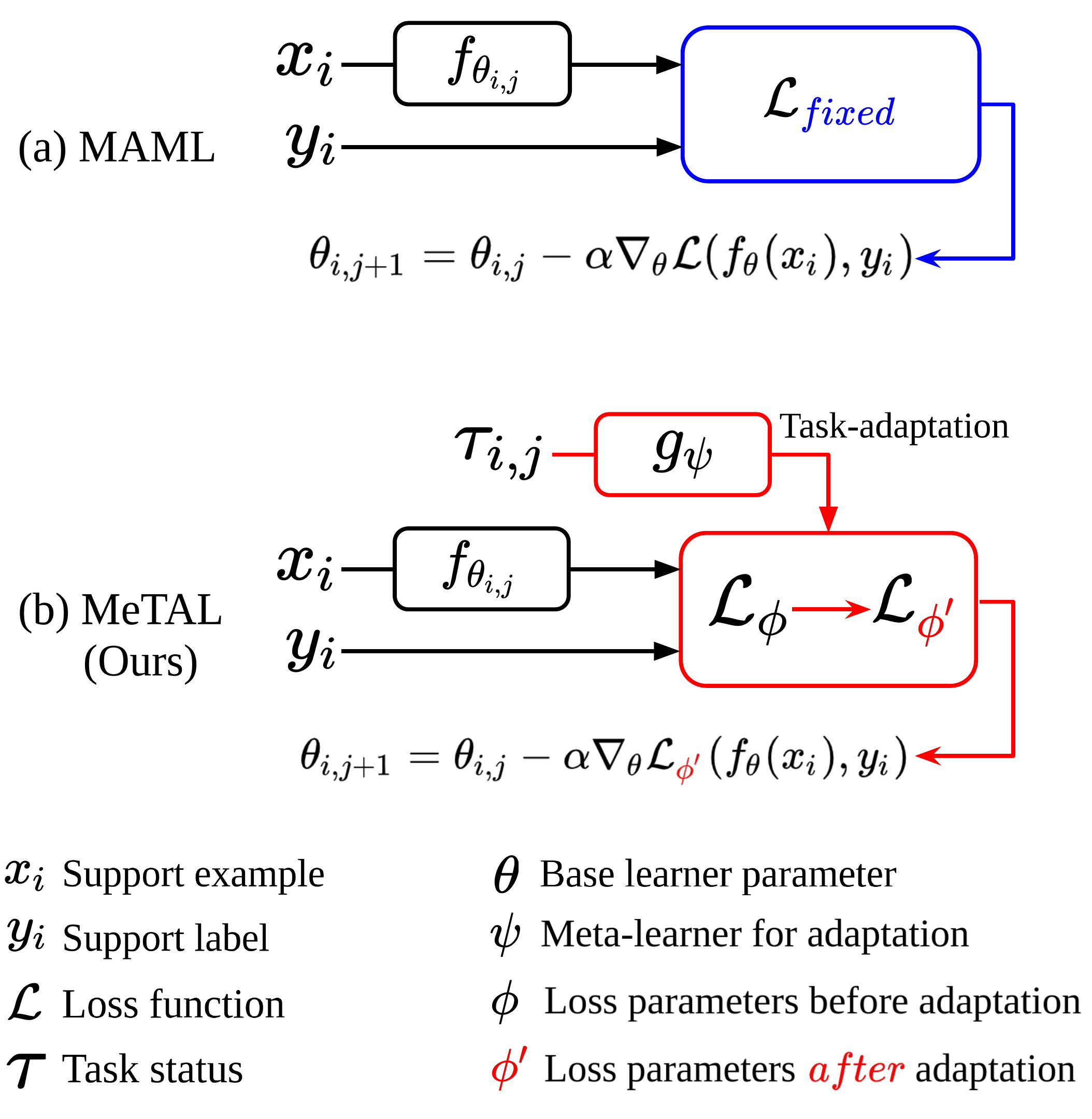}
	\caption{Overview of inner-loop optimization in optimization-based meta-learning frameworks.
	(a) Conventional approaches, such as MAML~\cite{finn2017model}, utilize a fixed given classical loss function (\eg cross entropy for classification) during adaptation to a task.
	(b) Our proposed scheme, MeTAL, instead meta-learns a loss function whose parameters $\bm{\phi}$ are made adaptive to the current task state $\bm{\tau}$ at the $j$-th step of adaptation to the $i$-th task.}
	\vspace{-1em}
	\label{fig:overview}
\end{figure}


\section{Introduction}
Training deep neural networks entails a tremendous amount of labeled data and the corresponding efforts, which hinder the prompt application to new domains.
As such, there has been growing interests in few-shot learning, in which the goal is to imbue the artificial intelligence systems with the capability of learning new concepts, given only few labeled examples (\eg \textit{support} examples).
The core challenge in few-shot learning is to alleviate the susceptibility of deep neural networks to overfitting under few-data regime and achieve generalization on new examples (\eg \textit{query} examples).

Recently, meta-learning~\cite{schmidhuber1987evolutionary,thrun2012learning}, \textit{a.k.a.} learning-to-learn, has emerged as one of the prominent methods for few-shot learning.
Meta-learning is used in the field of few-shot learning to learn a learning framework that can adapt to novel tasks and generalize under few-data regime.
Among the meta-learning algorithms, optimization-based meta-learning has enjoyed the attention from different domains for its flexibility that enables application across diverse domains.
Optimization-based meta-learning algorithms are often formulated as bi-level optimization~\cite{finn2017model,nichol2018first,ravi2017optimization}.
In such formulation, an outer-loop optimization trains a learning algorithm to achieve generalization while an inner-loop optimization uses the learning algorithm to adapt a base learner to a new task with few examples.

Model-agnostic meta-learning (MAML)~\cite{finn2017model}, one of the seminal optimization-based meta-learning methods, learns an initial set of values of network weights to achieve generalization.
The learned initialization serves as a good starting point for adapting to new tasks with few examples and few updates.
Although the learned initialization is trained to be a good starting point, MAML often faces the difficulty to achieve generalization, especially when tasks are diverse or significantly different between training and test phases~\cite{chen2019closerfewshot}.
Several works attempted to overcome the difficulty either by attempting to find a better initialization~\cite{baik2020learning,grant2018recasting,finn2018probabilistic,jamal2019task,vuorio2019multimodal,yao2019hierarchically} or a better fast adaptation process (inner-loop update rule)~\cite{antoniou2019how,behl2019alpha,lee2017gradient,Li2017meta,rusu2019meta}.
However, these methods resort to a simple loss function (\eg cross-entropy in classification) in the inner-loop optimization even though other auxiliary loss functions, such as $\ell_2$ regularization terms, can help achieve better generalization~\cite{baik2020meta}.

On the other hand, we focus on designing a better loss function for the inner-loop optimization in MAML framework.
As outlined in Figure~\ref{fig:overview}, we propose a new framework called \textbf{Me}ta-Learning with \textbf{T}ask-\textbf{A}daptive \textbf{L}oss Function (\textbf{MeTAL}) to learn an adaptive loss function that leads to better generalization for each task.
Specifically, MeTAL learns a task-adaptive loss function via two meta-learners: one meta-learner for learning a loss function and one meta-learner for generating parameters that transform a learned loss function.
Our task-adaptive loss function is designed to be flexible in that both labeled (\eg support) and unlabeled (\eg query) examples can be used together to adapt a base learner to each task during the inner-loop optimization.

The experimental results demonstrate that MeTAL greatly improves the generalization of MAML.
Owing to the simplicity and flexibility of MeTAL, we further demonstrate its effectiveness not only across different domains but also other MAML-based algorithms. 
When applied to other MAML-based algorithms, MeTAL consistently brings a substantial boost in generalization performance, introducing a new state-of-the-art performance among MAML-based algorithms.
This alludes to the significance of a task-adaptive loss function, which has drawn less attention in contrast to initialization schemes and inner-loop update rules. 
Overall experimental results underline that learning a better loss function for the inner loop optimization is important complementary component to learning a better inner-loop update or a better initialization.

\section{Related Work}
Few-shot learning aims to address scenarios where only few examples are available for each task. 
The ultimate goal is to learn new tasks with these given few examples while achieving generalization over unseen examples.
To this end, meta-learning algorithms attempt to tackle the few-shot learning problem via the learning of prior knowledge from previous tasks, which is then used to adapt to new tasks without overfitting~\cite{bengio1992optimization,hochreiter2001learning,schmidhuber1987evolutionary,schmidhuber1992learning,thrun2012learning}.

Depending on how the learning of prior knowledge and task adaptation process are formulated, meta-learning systems can generally be classified into metric-learning-based, black-box or network-based, and optimization-based approaches.  
Metric-learning-based approaches encode prior knowledge into an embedding space where similar (different) classes are closer (further apart) from each other~\cite{liu2019learning,koch2015siamese,NIPS2017_6996,Sung_2018_CVPR,vinyals2016matching}.
Black-box or network-based approaches employ a network or external memory to directly generates weights~\cite{munkhdalai2017meta,munkhdalai2018rapid}, weight updates~\cite{andrychowicz2016learning,hochreiter2001learning,ravi2017optimization}, or predictions~\cite{mishra2018simple,santoro2016meta}. 
Meanwhile, optimization-based approaches employ bi-level optimization to learn the learning procedures, such as initialization and weight updates, that will be used to adapt to new tasks with few examples~\cite{antoniou2019how,baik2020meta,baik2020learning,finn2017model,Li2017meta,nichol2018first,simon2020on}.

In this work, we focus on model-agnostic meta-learning (MAML) algorithm~\cite{finn2017model}, one of the most popular instances of the optimization approaches, owing to its simplicity and applicability across diverse problem domains.
MAML formulates prior knowledge as a learnable initialization, from which good generalization performance can be achieved for a new task after gradient-based fine-tuning with the given few examples.
While MAML is known for its simplicity and flexibility, it is also known for its relatively low generalization performance.
There has been recent studies on improving the overall performance either by enhancing the learning scheme of initialization~\cite{baik2020learning,jamal2019task,rusu2019meta,triantafillou2018meta,vuorio2019multimodal} or improving gradient-based fine-tuning process~\cite{antoniou2019how,baik2020meta,Li2017meta,simon2020on}. 

However, the aforementioned works still employ only a common loss function corresponding to a task (\textit{e.g.} cross-entropy in classification) during the inner-loop optimization.
Common deep learning frameworks, on the other hand, often use auxiliary loss terms, such as $\ell_2$ regularization term, in an effort to prevent overfitting.
As the goal of few-shot learning is to achieve generalization over unseen examples after adaptation with only few examples, using auxiliary loss terms seems like a natural choice. 
Few recently introduced methods have applied auxiliary loss functions in the inner-loop optimization to reduce the computational cost~\cite{lee2019meta,rajeswaran2019meta} or improve the generalization~\cite{goldblum2020unraveling}.
Other works attempted to learn loss functions for reinforcement learning (RL)~\cite{kirsch2020improving,oh2020discovering,sung2017learning,xu2020meta,xu2018meta,yu2018towards}, supervised learning~\cite{bechtle2021meta}, and incorporating unsupervised learning into few-shot learning~\cite{antoniou2019learning}.
However, loss functions from these methods either have task-specific requirements, such as environment interaction in RL, or remain fixed after training.
Fixed loss functions may be disadvantageous when new tasks may prefer different loss functions, especially in case train tasks and new tasks are significantly different (\eg cross-domain few-shot classification~\cite{chen2019closerfewshot}).

To this end, we propose a new meta-learning framework with a task-adaptive loss function (MeTAL).
In particular, a task-specific loss function is learned by a meta-network whose parameters are adapted to the given task.
Not only does MeTAL achieve outstanding performance, but it also maintains the simplicity and can be used jointly with other meta-learning algorithms. 

\section{Proposed Method}
\subsection{Preliminaries}
\subsubsection{Problem formulation}
We first introduce the preliminaries on meta-learning in the context of few-shot learning.
The meta-learning framework assumes a collection of tasks $\{\mathcal{T}_i\}^{T}_{i=1}$, each of which is assumed to be drawn from a task distribution $p(\mathcal{T})$.
Each task $\mathcal{T}_i$ consists of two disjoint sets of a dataset $\mathcal{D}_i$: a support set $\mathcal{D}^S_i$ and a query set $\mathcal{D}^Q_i$.
Each set, in turn, consists of a number of pairs of input $\bm{x}$ and output $\bm{y}$: $\mathcal{D}^S_i = \{(\bm{x}_i^s, \bm{y}_i^s)\}_{s=1}^{K}$ and $\mathcal{D}^Q_i = \{(\bm{x}_i^q, \bm{y}_i^q)\}_{q=1}^{M}$.

\begin{algorithm}[t]
\caption{Meta-learning with task-adaptive loss}
\label{algo:meta-training}
\begin{algorithmic}[1]
\REQUIRE Task distribution $p(\mathcal{T})$
\REQUIRE Learning rates $\alpha, \eta$
\REQUIRE Base learner network $f$, meta-networks $g$, $l$
\STATE Randomly initialize $\bm{\theta}, \bm{\phi}, \bm{\psi}$
\WHILE{not converged}
	\STATE Sample a batch of tasks $\mathcal{T}_i \sim p(\mathcal{T})$
	\FOR {each task $\mathcal{T}_i$}    
		\STATE Sample support set $\mathcal{D}^S_i = \{(\bm{x}_i^s, \bm{y}_i^s)\}_{s=1}^{K}$ from $\mathcal{T}_i$
		\STATE Sample query set $\mathcal{D}^Q_i = \{(\bm{x}_i^q, \bm{y}_i^q)\}_{q=1}^{M}$ from $\mathcal{T}_i$
		\STATE Initialize $\bm{\theta}_{i,0} = \bm{\theta}$
		\FOR {$j$ in number of inner-loop updates $J$}
		    \STATE Adapt $\bm{\theta}_{i,j+1} \leftarrow \bm{\theta}_{i,j}$ using Algorithm \ref{algo:inner-loop} \\
		\ENDFOR
		\STATE Compute the loss on the query set:\\ $\mathcal{L}(\mathcal{D}^Q_i;\bm{\theta}_{i,J})=\mathcal{L}(f(\bm{x}_i^q;\bm{\theta}_{i,J}),\bm{y}_i^q)$
	\ENDFOR
	\STATE Perform gradient descent to update weights: \\
	\small $(\bm{\theta}, \bm{\phi}, \bm{\psi}) \leftarrow (\bm{\theta},  \bm{\phi}, \bm{\psi}) - \eta \nabla_{(\bm{\theta}, \bm{\phi}, \bm{\psi})}\sum_{\mathcal{T}_i}\mathcal{L}(\mathcal{D}^Q_i;\bm{\theta}_{i,J})$
\ENDWHILE
\end{algorithmic}
\end{algorithm}

The goal of meta-learning is to learn a learning algorithm (formulated by a model with parameters $\bm{\phi}$) that can quickly learn tasks from a task distribution $p(\mathcal{T})$.
The learned learning algorithm is then leveraged to learn a new task $\mathcal{T}_i$ by adapting a base learner, parameterized by $\bm{\theta}$, using the task support examples $\mathcal{D}^S_i$, which is given by:
\begin{equation}\label{eq:task_adaptation}
    \bm{\theta}_i = \argmin_{\bm{\theta}}\mathcal{L}(\mathcal{D}^S_i;\bm{\theta},\bm{\phi}),
\end{equation}
where $\mathcal{L}$ denotes a loss function that evaluates the performance on a task.
As the support set $\mathcal{D}^S_i$ is used to learn a task, few-shot learning is often called $k$-shot learning when $k$ number of support examples is available for each task ($|\mathcal{D}^S_i| = K = k$).

The resultant task-specific base learner is represented by parameters $\bm{\theta}_i$.
The learned learning algorithm parameterized by $\bm{\phi}$ is then evaluated on how a task-specific base learner $\bm{\theta}_i$ generalizes to unseen query examples $\mathcal{D}^Q_i$.
Thus, the objective of the meta-learning algorithm becomes,
\begin{equation}\label{eq:meta_objective}
    \bm{\phi}^* = \argmin_{\bm{\phi}}\E_{\mathcal{T}_i \sim p(\mathcal{T})}[\mathcal{L}(\mathcal{D}^Q_i;\bm{\theta}_i,\bm{\phi})].
\end{equation}
\vspace{-2em}
\subsubsection{Model-agnostic meta-learning}
MAML~\cite{finn2017model} encodes prior knowledge into a learnable initialization that serves as a good initial set of values for weights of a base learner network across tasks.
This formulation, in which meta-learning initialization for a base learner, leads to bi-level optimization: inner-loop optimization and outer-loop optimization.
For the inner-loop optimization, a base learner is fine-tuned with support examples $\mathcal{D}^S_i$, from the learnable initialization $\bm{\theta}$, to each task for a fixed number of weight updates via gradient descent.
Thus, after initializing $\bm{\theta}_{i,0} = \bm{\theta}$, the task adaptation objective (Equation~\eqref{eq:task_adaptation}) is minimized via gradient-descent.
The inner-loop optimization at $j$-th step is denoted as:
\begin{equation}
\bm{\theta}_{i,j+1} = \bm{\theta}_{i,j} - \alpha\nabla_{\bm{\theta}_{i,j}} \mathcal{L}(\mathcal{D}^S_i;\bm{\theta}_{i,j}).
\label{eq:inner_update}
\end{equation}
Then, after $J$ number of inner-loop update steps, the task-specific base learner parameters $\bm{\theta}_i$ become $\bm{\theta}_{i,J}$.

\begin{algorithm}[t]
\caption{Inner-loop update subroutine}
\label{algo:inner-loop}
\begin{algorithmic}[1]
\REQUIRE Weights $\bm{\theta}_{i,j}$ of the base learner $f$
\REQUIRE Meta-networks $g$, $l$ with parameters $\bm{\phi}$
\REQUIRE Support set examples $\mathcal{D}^S_i$
\REQUIRE Unlabeled query set examples $\{\bm{x}_i^q\}_{q=1}^{M}$ \textbf{if} \textit{semi-supervised} setting
\STATE Compute the base learner output on the support set $f(\bm{x}^{s=1:K}_i;\bm{\theta}_{i,j})$ = $\{f(\bm{x}_i^s;\bm{\theta}_{i,j})\}^K_{s=1}$
\STATE Compute the loss on the support set:\\
$\mathcal{L}(\mathcal{D}^S_i;\bm{\theta}_{i,j})=\mathcal{L}(f(\bm{x}_i^{s=1:K};\bm{\theta}_{i,j}),\bm{y}_i^s)$
\STATE Compute the task state: \\
\IF {\textit{supervised}}
    \STATE $\bm{\tau}_{i,j}=[\mathcal{L}(\mathcal{D}^S_i;\bm{\theta}_{i,j}), \bm{\theta}_{i,j}, \{f(\bm{x}_i^{s=1:K};\bm{\theta}_{i,j})\}^K_{s=1}]$
\ELSIF{\textit{semi-supervised}}
     \STATE Compute the base learner output on the query set $f(\bm{x}^{q=1:M}_i;\bm{\theta}_{i,j})$ = $\{f(\bm{x}_i^q;\bm{\theta}_{i,j})\}^M_{q=1}$
    \STATE $\bm{\tau}_{i,j}=[\mathcal{L}(\mathcal{D}^S_i;\bm{\theta}_{i,j}), \bm{\theta}_{i,j}, $\\
    $f(\bm{x}_i^{s=1:K};\bm{\theta}_{i,j}), f(\bm{x}_i^{q=1:M};\bm{\theta}_{i,j})]$
\ENDIF
\STATE Compute the affine transformation parameters:\\ 
$\bm{\gamma}_{i,j}, \bm{\beta}_{i,j} = g(\bm{\tau}_{i,j}; \bm{\psi})$
\STATE Adapt the loss function parameters: \\
$ \bm{\phi}'_{i,j} = \bm{\gamma}_{i,j}\bm{\phi} + \bm{\beta}_{i,j}$ 
\STATE Compute task-adaptive loss: $\mathcal{L}_{\bm{\phi}'_{i,j}}(\bm{\tau}_{i,j})$
\STATE Perform gradient descent to adapt $f$ to $\mathcal{T}_i$: \\
$\bm{\theta}_{i,j+1} = \bm{\theta}_{i,j} - \alpha\nabla_{\bm{\theta}_{i,j}}\mathcal{L}_{\bm{\phi}'_{i,j}}(\bm{\tau}_{i,j})$
\end{algorithmic}
\end{algorithm}

In the case of the outer-loop optimization, the meta-learned initialization $\bm{\theta}$ is evaluated by the generalization performance of the task-specific base learner with parameters $\bm{\theta}_i$ (or $\bm{\theta}_{i,J}$) on unseen query examples $\mathcal{D}^Q_i$.
The evaluated generalization on unseen examples is then used as a feedback signal to update the initialization $\bm{\theta}$.
In other words, MAML minimizes the objective of the meta-learning algorithm, as in Equation~\eqref{eq:meta_objective}, as follows:
\begin{equation}
\bm{\theta}  \leftarrow \bm{\theta} - \eta \nabla_{\bm{\theta}} \sum_{\mathcal{T}_i}{\mathcal{L}(\mathcal{D}^Q_i;\bm{\theta}_i)}.
\label{eq:outer_update}
\end{equation}

\subsection{Meta-learning with Task-Adaptive Loss Function (MeTAL)}
\subsubsection{Overview}
Previous meta-learning formulations assume a fully-supervised setting for a given task $\mathcal{T}_i$ where they use labeled examples in the support set $\mathcal{D}^S_i$ to find the task-specific base learner $\bm{\theta}_i$ through minimizing a fixed given loss function $\mathcal{L}$.
On the other hand, we aim to control or meta-learn a loss function itself that would regulate the whole adaptation or inner-loop optimization process for better generalization.
We start with meta-learning an inner-loop optimization loss function $\mathcal{L}_{\bm{\phi}}(\cdot)$, modeled by a small neural network with its meta-learnable parameters $\bm{\phi}$. 
Thus, the inner-loop update in Equation \eqref{eq:inner_update} becomes,
\begin{equation}
\bm{\theta}_{i,j+1} = \bm{\theta}_{i,j} - \alpha\nabla_{\bm{\theta}_{i,j}}\mathcal{L}_{\bm{\phi}}(\bm{\tau}_{i,j}),
\label{eq:adaptive_inner_update}
\end{equation}
where $\bm{\tau}_{i,j}$ denotes the task state for $\mathcal{T}_i$ at time-step $j$, which is usually just the support set $\mathcal{D}^S_i$ in the case of the typical meta-learning formulation, as in Equation~\eqref{eq:inner_update}.
As different tasks (especially under cross-domain scenarios~\cite{chen2019closerfewshot}) may prefer different regularization or auxiliary loss functions or even loss functions itself during adaptation to achieve better generalization, we aim to learn to adapt a loss function itself to each task.
To enable a meta-learned loss function to be adaptive, one natural design choice could be to perform gradient descent, similar to how base learner parameters $\theta_i$ are updated as in Equation~\eqref{eq:inner_update}.
However, such design would result in a large computation graph, especially if a meta-learning algorithm is trained with higher-order gradients.
Alternatively, affine transformation could be applied to make a loss function adaptive to the given task.
Affine transformation conditioned on some input has been proved to be effective by several works in making feature responses adaptive~\cite{perez2018film,oreshkin2018tadam,jiang2019learning} and making meta-learned initialization adaptive~\cite{vuorio2019multimodal}.
In order to make a loss function task-adaptive without huge computation burden, we propose to dynamically transform loss function parameters $\bm{\phi}$ via affine transformation:
\begin{equation}
\bm{\phi}' = \bm{\gamma}\bm{\phi}+\bm{\beta},
\label{eq:loss_phi_affine}
\end{equation}
where $\bm{\phi}$ are the meta-learnable loss function parameters and $\bm{\gamma},\bm{\beta}$ are the transformation parameters generated by the meta-learner $g(\bm{\tau}_{j};\bm{\psi})$ which is parameterized by $\bm{\psi}$.

To train our meta-learning framework to generalize across different tasks, which involves optimizing the parameters $\bm{\theta}$, $\bm{\phi}$, and $\bm{\psi}$, the outer-loop optimization is performed with each task $\mathcal{T}_i$ given the respective task-specific learner $\bm{\theta}_{i}$ and its examples in the query set $\mathcal{D}^Q_i$ as in,
\begin{equation}
    (\bm{\theta}, \bm{\phi}, \bm{\psi}) \leftarrow (\bm{\theta},  \bm{\phi}, \bm{\psi}) - \eta \nabla_{(\bm{\theta}, \bm{\phi}, \bm{\psi})}\sum_{\mathcal{T}_i}\mathcal{L}(\mathcal{D}^Q_i;\bm{\theta}_{i}).
\label{eq:adaptive_outer_update}
\end{equation}
The overall training procedure of our method is summarized in Algorithm \ref{algo:meta-training}.
\vspace{-1em}
\subsubsection{Task-adaptive loss function}
Since our loss meta-learner $\mathcal{L}_{\bm{\phi}}$ and the meta-learner $g_{\bm{\psi}}$ are modeled using neural networks, their inputs can be formulated to contain auxiliary task-specific information on the intermediate learning status, which we define as a task state $\bm{\tau}$.
At $j$-th inner-loop step for a given task $\mathcal{T}_i$, in addition to classical loss information $\mathcal{L}(\mathcal{D}^S_i;\bm{\theta}_{i,j})$ (evaluated on the labeled support set examples $\mathcal{D}^S_i$), auxiliary learning state information, such as the network weights $\bm{\theta}_{i,j}$ and the output values $f(\bm{x}_i^s;\bm{\theta}_{i,j})$, can be included in the task state $\bm{\tau}_{i,j}$. 

Moreover, we can also include the base learner responses on the unlabeled examples $\bm{x}_i^q$ from the query set in the task state, which enables the inner-loop optimization to perform \textit{semi-supervised} learning.
This shows that our framework can use such additional task-specific information for fast adaptation, which was rarely utilized in previous MAML-based meta-learning algorithms, whereas metric-based meta-learning algorithms, such as~\cite{liu2019learning}, attempt to utilize unlabeled query examples to maximize the performance.
The semi-supervised inner-loop optimization maximizes the advantage of transductive setting (all query examples are assumed to be available at once), which MAML-based algorithms have implicitly used for better performance~\cite{nichol2018first}.
The procedure of inner-loop optimization with the task-adaptive loss function for both supervised and semi-supervised setting is organized in Algorithm \ref{algo:inner-loop}. 
\vspace{-1em}
\subsubsection{Architecture}
For our task-adaptive loss function $\mathcal{L}_{\bm{\phi}}$, we employ a 2-layer MLP with ReLU activation between the layers, which returns a single scalar value as output.
For the improved computational efficiency, the task state $\bm{\tau}_{i,j}$ used in the inner-loop optimization is formulated as a concatenation of mean of support set losses $\mathcal{L}(\mathcal{D}^S_i;\bm{\theta}_{i,j})$, layer-wise means of base learner weights $\bm{\theta}_{i,j}$, and example-wise mean of base learner output values $f(\bm{x}_i^s;\bm{\theta}_{i,j})$. 
Assuming a $L$-layer neural network for the base learner $f$ which returns $N$-dimensional output values (for $N$-way classification), the dimension of the task state $\bm{\tau}_{i,j}$ becomes {\small $1+L+N$}, which is computationally minimal. 
This can slightly increase under the semi-supervised learning setting where additional information can be derived from the responses of a base learner $f(\bm{x}_i^q;\bm{\theta}_{i,j})$ on the unlabeled query examples. 

Meta-network $g_{\bm{\psi}}$ also employs a 2-layer MLP with ReLU activation between the layers.
The network produces layer-wise affine transformation parameters that are applied to the loss function parameters $\bm{\phi}$. 
Since our meta-learning framework does not impose any constraint on the base learner $f$ and its target applications, our formulation is general and can be readily applied to any gradient-based differentiable learning algorithm.
For more details, please refer to the supplementary document and our code\footnote{The code is available at \url{https://github.com/baiksung/MeTAL}}.

\section{Experiments}
In this section, we perform experiments on several few-shot learning problems, such as few-shot classification, cross-domain few-shot classification, and few-shot regression to corroborate the effectiveness of task-adaptive loss functions.
All experimental results by our proposed method MeTAL are performed with semi-supervised inner-loop optimization, in which labeled support examples and unlabeled query examples are used together for the inner-loop optimization.
Note that we do not use extra data and that MeTAL simply takes more benefits from a transductive setting (all query examples are available at once) also employ for higher performance~\cite{nichol2018first}.

\subsection{Few-shot classification}
In few-shot classification, each task is defined as $N$-way $k$-shot classification, in which $N$ is the number of classes and $k$ is the number of examples (shots) per each class.

\begin{table*}[t]
\begin{center}
   \centering 
   \scalebox{1}{
   \begin{threeparttable}
   \begin{tabular}{lcc@{\hspace{-0.2cm}}ccc@{\hspace{-0.2cm}}ccc}
   \toprule[\heavyrulewidth]
          \multirow{2}{*}{\textbf{Model}}& \multirow{1}{*}{\textbf{Base learner}} & \phantom{abc}  &\multicolumn{2}{c}{\textbf{miniImageNet}} & \phantom{abc}& \multicolumn{2}{c}{\textbf{tieredImageNet}} \\
           \cmidrule{4-5}\cmidrule{7-8}
          &\textbf{Backbone}& &1-shot & 5-shot && 1-shot & 5-shot\\
   \midrule
   MAML + L2F ~\cite{baik2020learning} & 4-CONV & & $52.10 \pm 0.50\%$ & $69.38 \pm 0.46\%$& & $54.40 \pm 0.50\%$ & $73.34 \pm 0.44\%$ \\ 
   MAML~\cite{finn2017model} & 4-CONV && $48.70 \pm 1.75\%$ & $63.11 \pm 0.91\%$ && - & - \\
   MAML\tnote{\textdaggerdbl} & 4-CONV && $49.64\pm 0.31\%$ & $64.99 \pm 0.27\%$ &&$50.98\pm 0.26\%$ & $66.25 \pm 0.19\%$ \\
   \textbf{MeTAL (Ours)} & 4-CONV & &  $52.63 \pm 0.37\%$ & $70.52 \pm 0.29\%$ &&$54.34\pm 0.31\%$ & $70.40 \pm 0.21\%$ \\ \hdashline \\[-8pt]
   MAML++ + SCA~\cite{antoniou2019learning}& 4-CONV && $54.24 \pm 0.99\%$ & $71.85 \pm 0.53\%$ &&- & - \\
   MAML++~\cite{antoniou2019how}& 4-CONV && $52.15 \pm 0.26\%$ & $68.32 \pm 0.44\%$ &&- & -\\
   MAML++ + \textbf{MeTAL (Ours)}& 4-CONV & & $57.18 \pm 0.42\%$ & $72.89 \pm 0.44\%$ &&$59.93\pm 0.36\%$ & $75.39 \pm 0.29\%$ \\\hdashline \\[-8pt]
   ALFA + MAML ~\cite{baik2020meta} & 4-CONV & &$50.58 \pm 0.51\%$ & $69.12 \pm 0.47\%$ & &$53.16 \pm 0.49\%$ & $70.54 \pm 0.46\%$ \\
   ALFA + \textbf{MeTAL (Ours)} & 4-CONV & & $\textbf{57.75} \pm \textbf{0.38}\%$ & $\textbf{74.10} \pm \textbf{0.43}\%$ & &$\textbf{60.29} \pm \textbf{0.37}\%$ & $\textbf{75.88} \pm \textbf{0.29}\%$  \\
   \midrule
   MAML++~\cite{antoniou2019how}& DenseNet && $58.37 \pm 0.27\%$ & $75.50 \pm 0.19\%$ & &-& -\\
   SCA + MAML++~\cite{antoniou2019learning}& DenseNet && $62.86 \pm 0.79\%$ & $77.64 \pm 0.40\%$ & &-& -\\
   MAML + L2F ~\cite{baik2020learning} & ResNet12 & & $59.71 \pm 0.49\%$ & $77.04 \pm 0.42\%$ & &$64.04 \pm 0.48\%$ & $81.13 \pm 0.39\%$\\ \hdashline \\[-8pt]
   MAML\tnote{\textdaggerdbl} & ResNet12 & &$58.60 \pm 0.42\%$ & $69.54 \pm 0.38\%$ & &$59.82 \pm 0.41\%$ & $73.17 \pm 0.32\%$\\
   \textbf{MeTAL (Ours)} & ResNet12 & & $59.64 \pm 0.38\%$ & $76.20 \pm 0.19\%$ & &$63.89\pm 0.43\%$ & $80.14 \pm 0.40\%$\\ \hdashline \\[-8pt]
   ALFA + MAML ~\cite{baik2020meta} & ResNet12 & & $59.74 \pm 0.49\%$ & $77.96 \pm 0.41\%$&& $64.62 \pm 0.49\%$ & $82.48 \pm 0.38\%$ \\
   ALFA + \textbf{MeTAL (Ours)} & ResNet12 & &  $\textbf{66.61} \pm \textbf{0.28}\%$ & $\textbf{81.43} \pm \textbf{0.25}\%$ & & $\textbf{70.29} \pm \textbf{0.40}\%$ &$\textbf{86.17} \pm \textbf{0.35}\%$\\
   \midrule
   SNAIL~\cite{mishra2018simple} & ResNet12 && $55.71 \pm 0.99\%$ & $68.88 \pm 0.92\%$ && - & - \\
   AdaResNet~\cite{munkhdalai2018rapid} & ResNet12 && $56.88 \pm 0.62\%$ & $71.94 \pm 0.57\%$ && - & - \\
   TADAM~\cite{oreshkin2018tadam} & ResNet12 && $58.50 \pm 0.30\%$ & $76.70 \pm 0.30\%$ && - & - \\
   LEO-trainval$^*$~\cite{rusu2019meta} & WRN-28-10 & & $61.76 \pm 0.08\% $ & $77.59 \pm 0.12 \%$ && $66.33 \pm 0.05\%$ & $81.44 \pm 0.09\%$ \\
   MetaOpt \tnote{\textdagger} ~\cite{lee2019meta} & ResNet12 & & $62.64 \pm 0.61\% $ & $78.63 \pm 0.46 \%$ && $65.99 \pm 0.72\%$ & $81.56 \pm 0.53\%$ \\
   \bottomrule[\heavyrulewidth] 
   \end{tabular}
   \begin{tablenotes}
   \small\item[*] Pretrained
   \small\item[\textdagger] Trained with data augmentation.
   \small\item[\textdaggerdbl] Reproduced.
   \end{tablenotes}
   \end{threeparttable}
   }
      \caption{5-way 1-shot and 5-way 5-shot classification test accuracy on miniImageNet and tieredImageNet.}
   \vspace{-1em}
   \label{tab:result_mini_tiered}
\end{center}
\end{table*}
\vspace{-1em}
\subsubsection{Datasets}
The most commonly used datasets for few-shot classification are two ImageNet\cite{ILSVRC15}-derivative datasets: miniImageNet~\cite{ravi2017optimization} and tieredImageNet~\cite{ren2018meta}.
Both datasets are composed of three disjoint subsets (train, validation, and test sets), each of which consists of images with the size of $84 \times 84$.
The datasets differ in how classes are split into disjoint subsets.
miniImageNet randomly samples and groups classes into 64 classes for meta-training, 16 for meta-validation, and 20 for meta-test~\cite{ravi2017optimization}.
tieredImageNet, on the other hand, groups classes into 34 categories according to ImageNet class hierarchy and splits groups into 20 categories for meta-training, 6 for meta-validation, and 8 for meta-test~\cite{ren2018meta} in an effort to minimize class similarities between the three disjoint sets.
\vspace{-1em}
\subsubsection{Results}
We assess our method MeTAL and compare with other MAML variants on miniImageNet and tieredImageNet under two typical settings: 5-way 5-shot and 5-way 1-shot classification, as presented in Table~\ref{tab:result_mini_tiered}.
The results demonstrate that not only does MeTAL greatly improve the generalization performance of MAML but also can be applied in conjunction with other MAML variants, such as MAML++~\cite{antoniou2019how} and ALFA~\cite{baik2020meta}, to bring further improvement.
MAML++ learns fixed step-and-layer-wise inner-loop learning rates while ALFA learns task-adaptive inner-loop learning rates and regularization terms.
Although these methods do not consider a loss function to be learnable, if a loss function is regarded as a part of the model, then MeTAL may be seen as a more general extension of these methods.
However, the further improvement brought by MeTAL upon these methods demonstrates that improving the inner-loop optimization objective function is not just a simple extension but a complementary and orthogonal factor.
The major contribution of MeTAL lies in formulating an inner-loop loss function to be learnable and task-adaptive.

Furthermore, MeTAL, together with ALFA~\cite{baik2020meta}, greatly outperforms other models that either use larger networks, such as DenseNet or WideResNet, or are pretrained or trained with data augmentation.
These results suggest the effectiveness of our learned task-adaptive loss function in achieving better generalization.

\subsection{Cross-domain few-shot classification}
\begin{table}[t]
   \centering
   \scalebox{0.9}{
   \begin{tabular}{lcc}
   \toprule[\heavyrulewidth]
          & \textbf{Base learner} & \textbf{miniImageNet}\\
          & \textbf{Backbone} &  $\rightarrow$ \textbf{CUB}\\
   \midrule
   MAML & 4-CONV & $52.70 \pm 0.32\%$\\ 
   \textbf{MeTAL (Ours)} & 4-CONV & $\textbf{58.20} \pm \textbf{0.24}\%$\\  \hdashline \\[-8pt]
   ALFA + MAML & 4-CONV & $58.35 \pm 0.25\%$\\
   ALFA + \textbf{MeTAL (Ours)} & 4-CONV & $\textbf{66.37} \pm \textbf{0.17}\%$ \\
   \midrule
   MAML & ResNet12 & $53.83 \pm 0.32\%$\\
   \textbf{MeTAL (Ours)} & ResNet12 & $\textbf{61.29} \pm \textbf{0.21}\%$\\ \hdashline \\[-8pt]
   ALFA + MAML & ResNet12 &$63.64 \pm 0.42\%$\\
   ALFA + \textbf{MeTAL (Ours)} & ResNet12 & $\textbf{70.22} \pm \textbf{0.14}\%$\\
   \bottomrule[\heavyrulewidth] 
   \end{tabular}
   }
   \caption{5-way 5-shot cross-domain few-shot classification. Models are trained on miniImageNet and evaluated on CUB.} 
   \label{tab:cross_domain}
   \vspace{-1em}
\end{table}
Cross-domain few-shot classification, introduced by Chen~\etal~\cite{chen2019closerfewshot}, tackles a more challenging and practical few-shot classification scenario, where meta-train tasks and meta-test tasks are sampled from different task distributions.
Such scenario is intentionally designed to create a large domain gap between meta-train and meta-test, thereby assessing the susceptibility of meta-learning algorithms to meta-level overfitting.
Specifically, a meta-learning algorithm can be said to be meta-overfitted if the algorithm is too dependent on prior knowledge from previously seen meta-train tasks, instead of focusing on the given few examples to learn a new task.
This meta-level overfitting will result in a learning system being more likely to fail to adapt to a new task that is sampled from substantially different task distributions. 
\vspace{-1.2em}
\subsubsection{Datasets}
To simulate such challenging scenario, Chen~\etal~\cite{chen2019closerfewshot} first meta-trains algorithms on miniImageNet~\cite{ravi2017optimization} and evaluates them on CUB dataset (CUB-200-2011)~\cite{WahCUB_200_2011} during meta-test.
In contrast to ImageNet that is compiled for general classification tasks, CUB targets fine-grained classification.
Following the protocol from~\cite{chen2019closerfewshot}, 200 classes of the dataset are split into 100 meta-train, 50 meta-validation, and 50 meta-test sets.
\vspace{-1em}
\subsubsection{Results}
Table~\ref{tab:cross_domain} presents the performance of MAML~\cite{finn2017model}, one of recent MAML variants ALFA~\cite{baik2020meta}, and MeTAL when they are trained on miniImageNet meta-train set and evaluated on CUB meta-test set.
Similar to the few-shot classification results outlined in Table~\ref{tab:result_mini_tiered}, MeTAL is shown to greatly improve the generalization even under the more challenging cross-domain few-shot classification scenario.
In fact, MeTAL improves the performance of both MAML and ALFA + MAML by greater extent in cross-domain few-shot classification ($\sim8\%$) than in few-shot classification ($\sim4\%$).
This implies the effectiveness of MeTAL in learning new tasks from different domains and its robustness to the domain gap, stressing the importance of task-adaptive loss function.
Another observation can be made regarding the results: the increase in generalization performance made by MeTAL on ALFA + MAML is as great as on MAML, indicating the orthogonality of the problem MeTAL attempts to solve.
ALFA~\cite{baik2020meta} also aims to improve the inner-loop optimization but the difference is they focus on developing a new weight-update rule (gradient descent).
On the other hand, we focus on a loss function that is used in the inner-loop optimization.
The consistent generalization improvement by MeTAL across different baselines and architectures signifies that designing a better inner-loop optimization loss function is important factor and complementary to designing a better weight-update rule.

\subsection{Few-shot regression}
To demonstrate the flexibility and applicability of our method MeTAL, we evaluate MAML and MeTAL on few-shot regression, or $k$-shot regression.
In $k$-shot regression, each task is to estimate a given unknown function when only very few number ($k$) of whose sampled points are given. 
The task distribution consists of tasks that have a target function with parameters whose values vary within a defined range.
In this work, we follow the general settings Finn~\etal~\cite{finn2017model} have used for evaluating MAML.
Specifically, each task has a sinusoid  $y(x) = A sin(\omega x+b)$ as a target function whose parameter values are within the following ranges: amplitude $A \in [0.1, 5.0]$, frequency $\omega \in [0.8,1.2]$, and phase $b \in [0, \pi]$.
For each task, input data points $x$ are sampled from $[-5.0,5.0]$.
Regression is performed by performing a single gradient descent on a base learner whose neural architecture is composed of 3 layers of size 80 with ReLU non-linear activation functions in between.
The performance is measured in mean-square error (MSE) between the estimated output values $\hat{y}$ and ground-truth output values $y$.

Table~\ref{tab:regression} outlines the regression results from MAML~\cite{finn2017model} and MeTAL under $5$-shot, $10$-shot, and $20$-shot settings.
Again, MeTAL demonstrates the consistent performance improvement across different settings.
This proves the applicability and flexibility of the proposed task-adaptive loss function, learned by MeTAL.
\begin{table}[t]
   \scalebox{0.9}{
   \centering 
   
    \begin{tabular}{lccc}
        \toprule
        & \multicolumn{1}{c}{5 shots} & \multicolumn{1}{c}{10 shots} & \multicolumn{1}{c}{20 shots}\\
        \hline \\[-1.8ex] 
        MAML  & $0.86 \pm 0.23$ & $0.50 \pm 0.12$ & $0.26 \pm 0.08$ \\
        \textbf{MeTAL (Ours)} & $\textbf{0.74} \pm \textbf{0.18}$ & $\textbf{0.44} \pm \textbf{0.11}$ & $\textbf{0.21} \pm \textbf{0.06}$ \\  
        \bottomrule
\end{tabular}
}
\caption{$k$-shot regression: mean-square error (MSE) measured over 100 sampled points with 95\% confidence intervals.} 
\label{tab:regression}
\end{table}

\subsection{Ablation studies}
To investigate the contribution of each module in MeTAL, we perform ablation study experiments in this section.
In particular, we analyze the effectiveness of the task state information, the learning of a loss function, the task-adaptive loss function, and the semi-supervised inner-loop optimization formulation.
All ablation study experiments were performed with a base learner that has 4-CONV backbone under $5$-way $5$-shot few-shot classification.

\begin{table}[t]
   \centering 
   
    \begin{tabular}{@{}l|ccc@{}}
        \toprule
        & cross entropy & learned loss & accuracy\\\midrule
        \textbf{(1)}  & \cmark &        &  $64.99 \pm 0.27\%$ \\
        \textbf{(2)}  &        & \cmark &  $67.42 \pm 0.34\%$ \\
        \textbf{(3)}  & \cmark & \cmark &  $67.86 \pm 0.42\%$ \\
        \bottomrule
\end{tabular}
\caption{The study on the effectiveness of learning a loss function during the inner-loop optimization. Model \textbf{(1)} denotes MAML.} 
\vspace{-1em}
\label{tab:loss_function}
\end{table}
\vspace{-1em}
\subsubsection{Learning a loss function}
First, we analyze the importance of learning an inner-loop optimization loss function.
In detail, the performance is measured when the inner-loop optimization is performed with a loss function that is learned without the adaptation (\ie only a meta-network $\mathcal{L}_{\bm{\phi}}$ is used for model \textbf{(2)}, \textbf{(3)}) and compared with when a simple cross entropy is used (\ie MAML is represented as model \textbf{(1)}).

\begin{table}[t]
   \centering 
    \begin{tabular}{@{}l|ccc@{}}
        \toprule
        & Task-adaptive & semi-supervised & accuracy\\\midrule
        \textbf{(2)}  &        &        &  $67.42 \pm 0.34\%$ \\
        \textbf{(4)}  & \cmark &        &  $68.56 \pm 0.36\%$\\
        \textbf{(5)}  &        & \cmark &  $68.24 \pm 0.31\%$ \\
        \textbf{(6)}  & \cmark & \cmark & $\textbf{70.52} \pm \textbf{0.29}\%$ \\
        \bottomrule
    \end{tabular}
    \caption{The study on the effectiveness of task-adaptive loss function and semi-supervised setting during the inner-loop optimization. Model \textbf{(6)} corresponds to MeTAL. Model \textbf{(2)} refers to MAML trained with a fixed learned loss from Table~\ref{tab:loss_function}.}
    \label{tab:task_adaptive_semi}
\end{table}

\begin{table}[t]
   \centering 
    \begin{tabular}{@{}l|c@{\hspace{0.5\tabcolsep}}ccc@{}}
        \toprule
         & cross entropy &\multirow{2}{*}{weight} & \multirow{2}{*}{prediction} & \multirow{2}{*}{accuracy} \\
        &  (support set)  & &&\\\midrule
        \textbf{(7)}& \cmark  &        &        &  $67.86 \pm  0.41\%$ \\
        \textbf{(8)}& \cmark  & \cmark &        &  $68.66 \pm  0.46\%$\\
        \textbf{(9)}& \cmark  &        & \cmark &  $67.94 \pm 0.47\%$ \\
        \textbf{(6)}& \cmark  & \cmark & \cmark & $\textbf{70.52} \pm \textbf{0.29}\%$ \\
        \bottomrule
    \end{tabular}
    \caption{Investigation on the role of each factor of the task state $\bm{\tau}$.}
    \vspace{-1em}
    \label{tab:task_state}
\end{table}
The ablation study results summarized in Table~\ref{tab:loss_function} show that the learned a loss function helps MAML achieve better generalization, suggesting that a meta-learner has managed to learn loss function that is useful for generalization. 
Further, when cross entropy and a learned loss are used together, there is no significant difference from when only a learned loss is used, implying that a learned loss is able to maintain cross entropy loss information that is fed as input.
\vspace{-1.5em}
\subsubsection{Task-adaptive loss function}
We then examine the influence of task-adaptive loss function on the overall proposed framework.
To this end, we use a meta-model $g_{\bm{\psi}}$ to generate affine transformation parameters, which are then used to adapt the parameters of a loss function meta-network $\mathcal{L}_{\bm{\phi}}$ in model \textbf{(2)} from Table~\ref{tab:loss_function}, according to Equation~\eqref{eq:loss_phi_affine}.
The derived meta-learning algorithm, which is MeTAL without semi-supervised inner-loop optimization, is denoted as model \textbf{(4)} in Table~\ref{tab:task_adaptive_semi}.
As shown in the table, the meta-learning algorithm benefits from a task-adaptive learned loss function, compared to a fixed learned function.
\begin{figure*}[t!]
	\centering
	\includegraphics[width=0.45\linewidth]{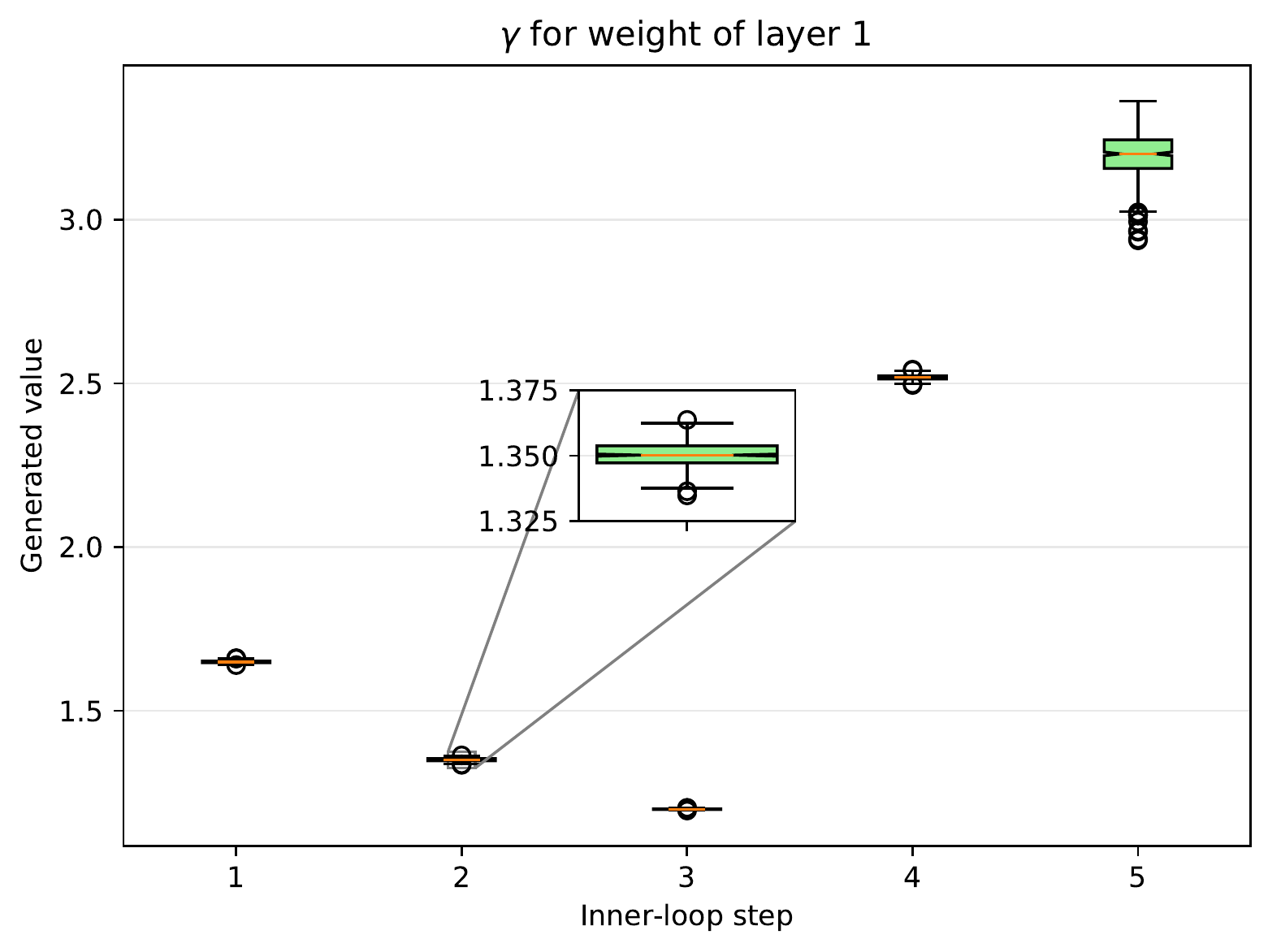}~~~~~
	\includegraphics[width=0.45\linewidth]{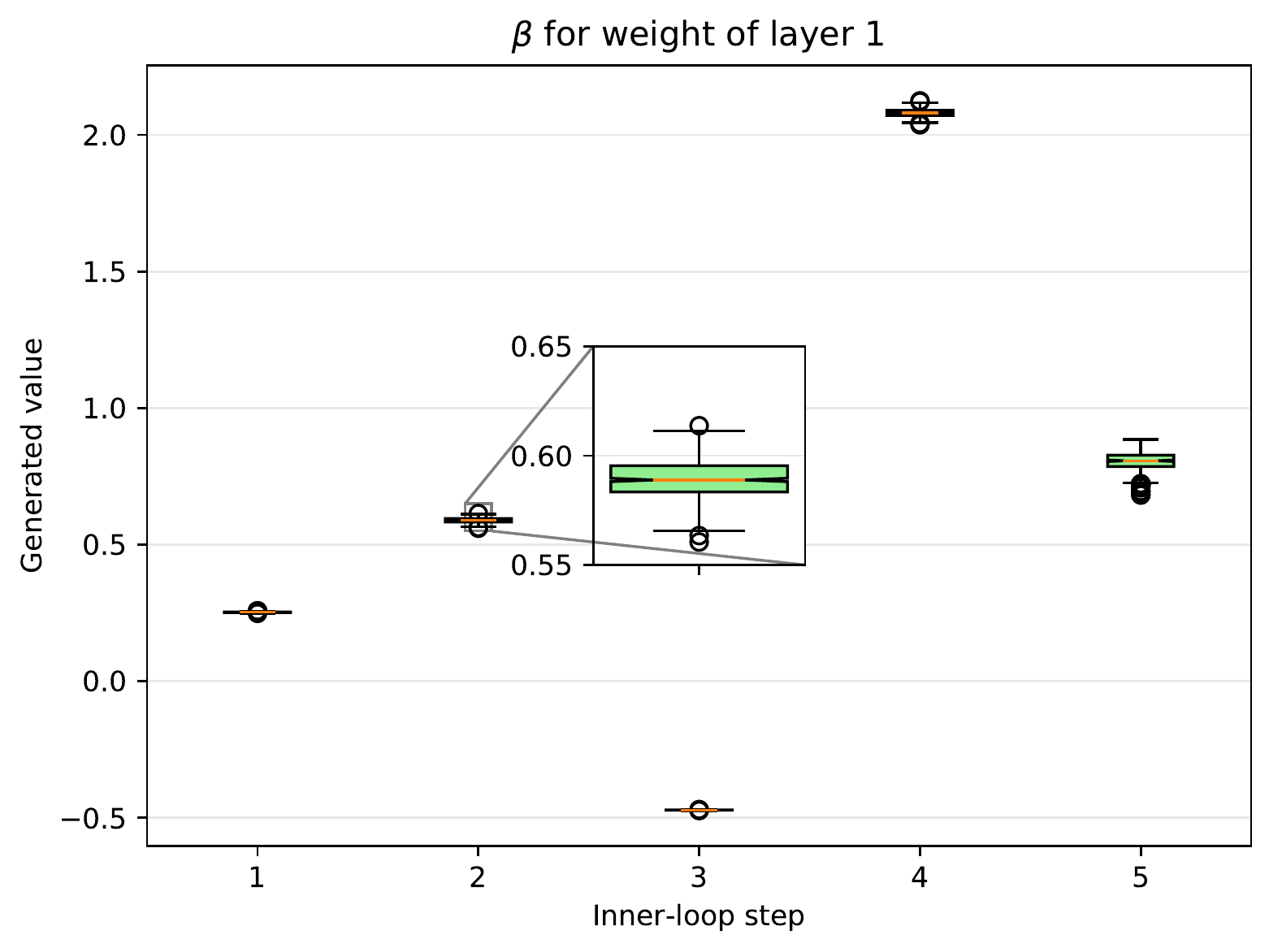}
	\\
	 \caption{
	 The illustration of the generated affine transformation parameters, $\gamma$ and $\beta$, that are generated by one of our proposed meta-networks, $g$. 
	 These values are then used to adapt the loss meta-network $l$ to the given task.
	 In particular, the figure visualizes the generated values for the weight of the first layer of the loss meta-network $l$. 
	 The generated values demonstrate its dynamic range across the inner-loop steps, which indicates a different loss function is preferred at each step.
	 Additionally, different values are observed among different tasks, especially at the last inner-loop step, alluding to diverse preferences of tasks on the inner-loop optimization loss functions.
	 } 
	 \vspace{-1em}
	\label{fig:viz}
\end{figure*}
\vspace{-1.5em}
\subsubsection{Semi-supervised inner-loop optimization}
Next, we look into the effectiveness of the semi-supervised inner-loop optimization formulation.
Similar to the task-adaptive loss function ablation study, we first derive a new model that is created by adding the semi-supervised inner-loop optimization formulation (using labeled support examples and unlabeled query examples together for fast adaptation via learned loss function) to model \textbf{(2)} from Table~\ref{tab:loss_function}.
Consequently, the resulting model, which is denoted as model \textbf{(5)}, lacks the task-adaptive property, compared to our final method MeTAL.
While the semi-supervised inner-loop optimization contributes to the performance improvement, it still lags behind the full algorithm MeTAL (denoted as model \textbf{(6)}), alluding to the significance of the task-adaptive loss function. 
\vspace{-1em}
\subsubsection{Task state}

We perform another ablation study for the investigation on the effect of each factor of the task state $\bm{\tau}$: namely, the current weight values of base learner $\bm{\theta}_{i,j}$, the output of network ($f(\bm{x}_i^s;\bm{\theta}_{i,j})$ for support and $f(\bm{x}_i^q;\bm{\theta}_{i,j})$ for query), and cross entropy loss for support set $\mathcal{L}(\mathcal{D}^S_i;\bm{\theta}_{i,j})=\mathcal{L}(f(\bm{x}_i^s;\bm{\theta}_i),\bm{y}_i^s)$.
The ablation results are summarized in Table~\ref{tab:task_state}.
When the original cross entropy on support examples is not included in the task state, the whole inner-loop optimization becomes unsupervised learning setting as no ground-truth information is involved during the inner-loop optimization.
In such case, as one would expect, MeTAL struggles to achieve generalization, thereby excluding these results in the table.
When conditioned on cross entropy loss as a task state (model \textbf{(7)}), MeTAL manages to bring better generalization.
Furthermore, including weight (model \textbf{(8)}) or prediction (model \textbf{(9)}) into task state contributes to further improvement.
Finally, when all factors of the task state are used, MeTAL achieves the best performance, emphasizing the importance of each factor.

\subsection{Visualization}

Figure~\ref{fig:viz} illustrates the affine transformation parameters $\gamma$ and $\beta$ generated by one of our proposed meta-networks $g$ across tasks (represented as boxplot) for each inner-loop step.
Observing how the generated $\gamma$ and $\beta$ values vary across the inner-loop steps, we can claim that MeTAL manages to dynamically adapt a loss function as the learning state changes during the inner-loop optimization.
Further, the generated parameter values are shown to vary among tasks, especially at the last inner-loop step.
This may suggest that the overall framework is trained to make the most difference among tasks at the last step.
Regardless, the dynamic range of the generated affine transformation parameter values among tasks validates the effectiveness of the MeTAL in adapting a loss function to the given task. 


\section{Conclusion}
In this work, we propose a meta-learning framework with a task-adaptive loss function for few-shot learning.
The proposed scheme, named MeTAL, learns a loss function that adapts to each task based on the current task state during the inner-loop optimization.
Consequently, MeTAL is able to learn a loss function that is specifically needed by each task for better generalization. 
Furthermore, not only does the flexibility of MeTAL enable it to be applied to different MAML variants and problem domains, but also allows for the semi-supervised inner-loop optimization, in which labeled support examples and unlabeled query examples are used jointly for adapting to tasks.
Overall, the experimental results underline the importance of learning a good loss function for each task, which has drawn relatively less attention, compared to a weight-update rule or initialization in the context of few-shot learning.

\vspace{-1em}
\paragraph{Acknowledgment}
\noindent This work was supported in part by IITP grant funded by the Korea government [No. 2021-0-01343, Artificial Intelligence Graduate School Program (Seoul National University)], and in part by AIRS Company in Hyundai Motor and Kia through HMC/KIA-SNU AI Consortium Fund.

{\small
\bibliographystyle{ieee_fullname}
\bibliography{main}
}

\clearpage

\includepdf[pages=1]{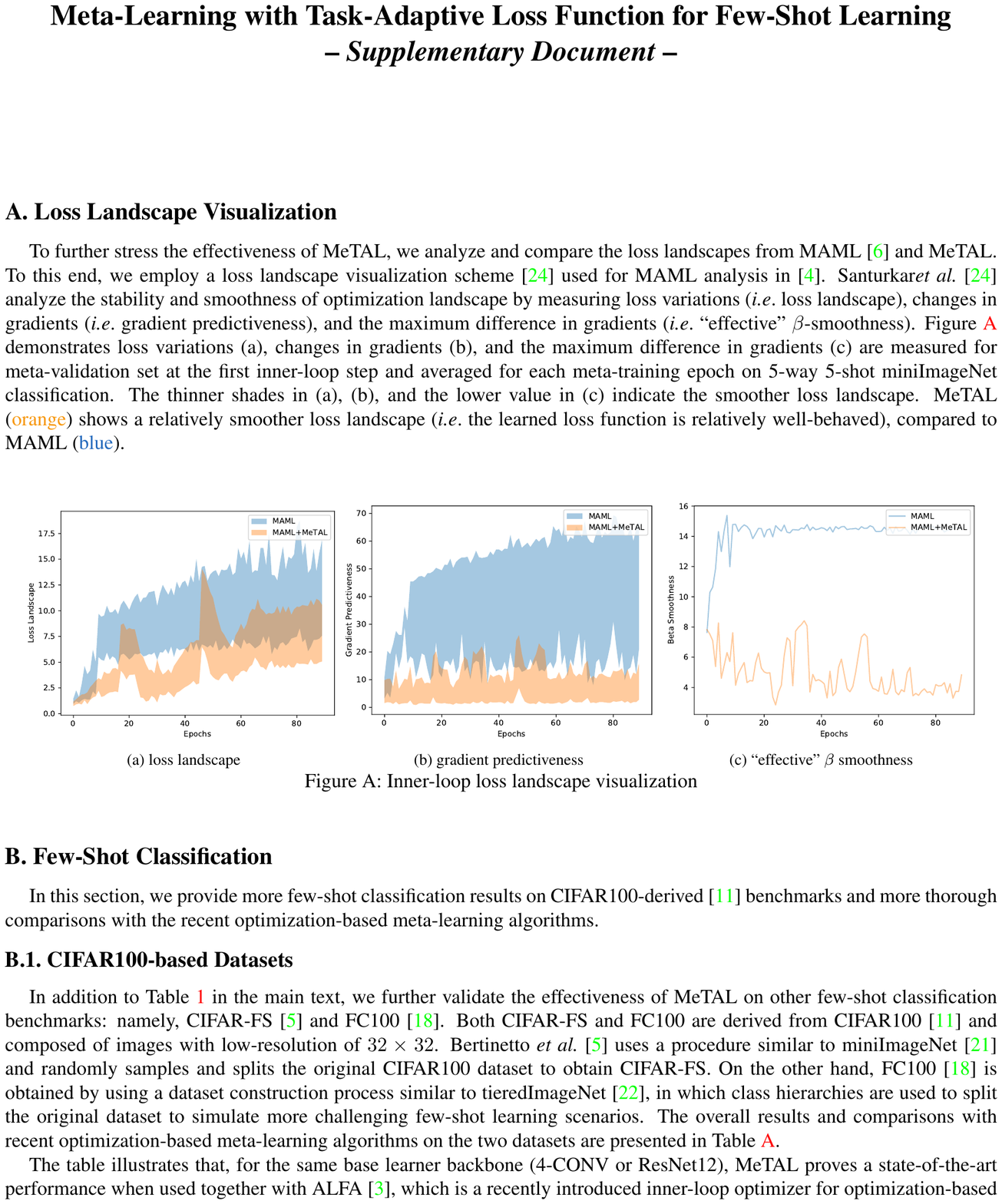}
\includepdf[pages=2]{supp.pdf}
\includepdf[pages=3]{supp.pdf}
\includepdf[pages=4]{supp.pdf}
\includepdf[pages=5]{supp.pdf}
\includepdf[pages=6]{supp.pdf}
\includepdf[pages=7]{supp.pdf}

\end{document}


\onecolumn
\title{Meta-Learning with Task-Adaptive Loss Function for Few-Shot Learning\\
-- \textit{Supplementary Document} -- }

\maketitle

\ificcvfinal\thispagestyle{empty}\fi

\section{Loss Landscape Visualization}
\begin{figure*}[h!]
\begin{center}
\subfloat[loss landscape]{
    \includegraphics[width=0.32\linewidth]{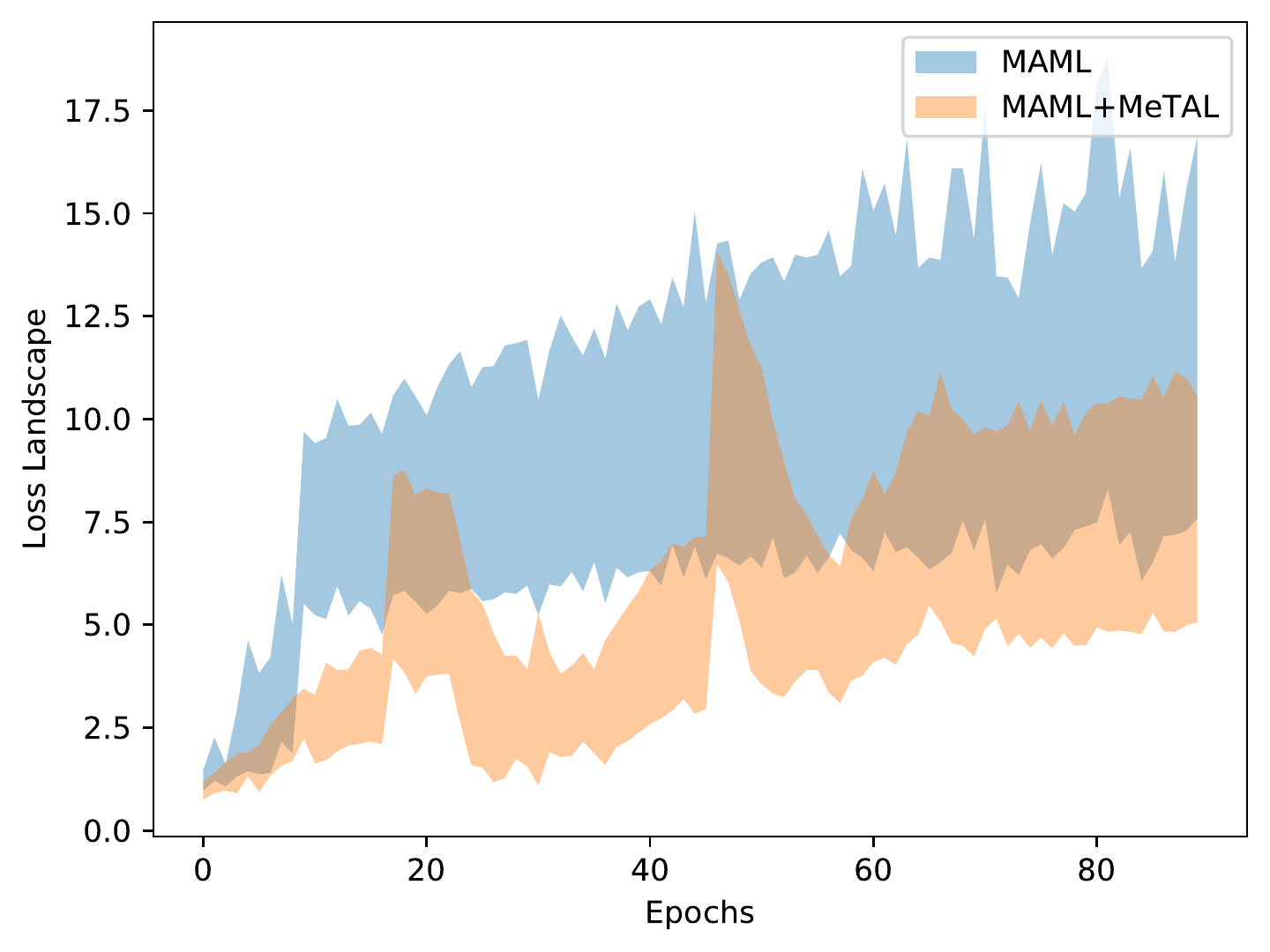}
    \label{fig:loss}
}
\hspace{-1em}
\subfloat[gradient predictiveness]{
    \includegraphics[width=0.32\linewidth]{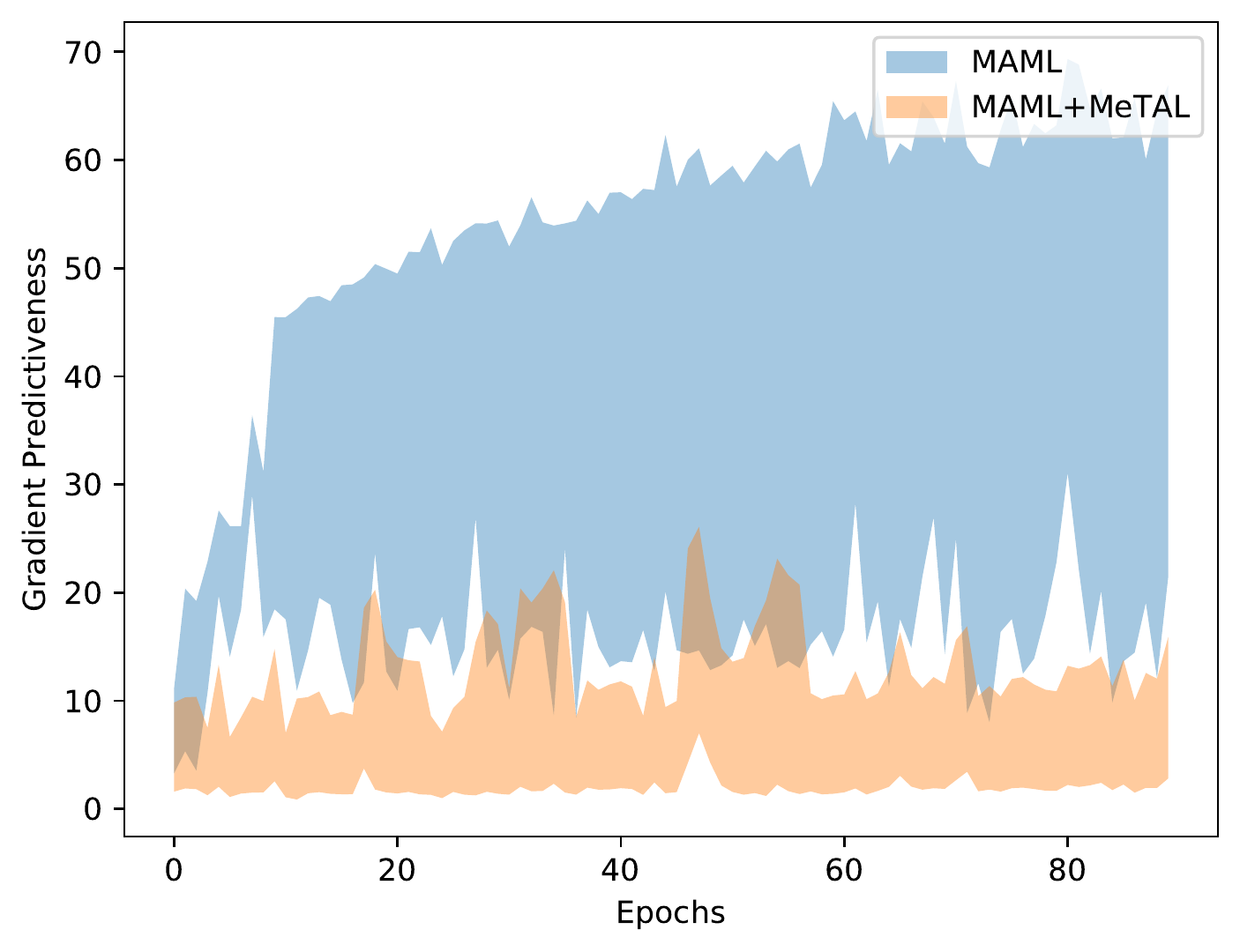}
    \label{fig:gradient}
}
\hspace{-0.5em}
\subfloat[``effective'' $\beta$ smoothness]{
    \includegraphics[width=0.32\linewidth]{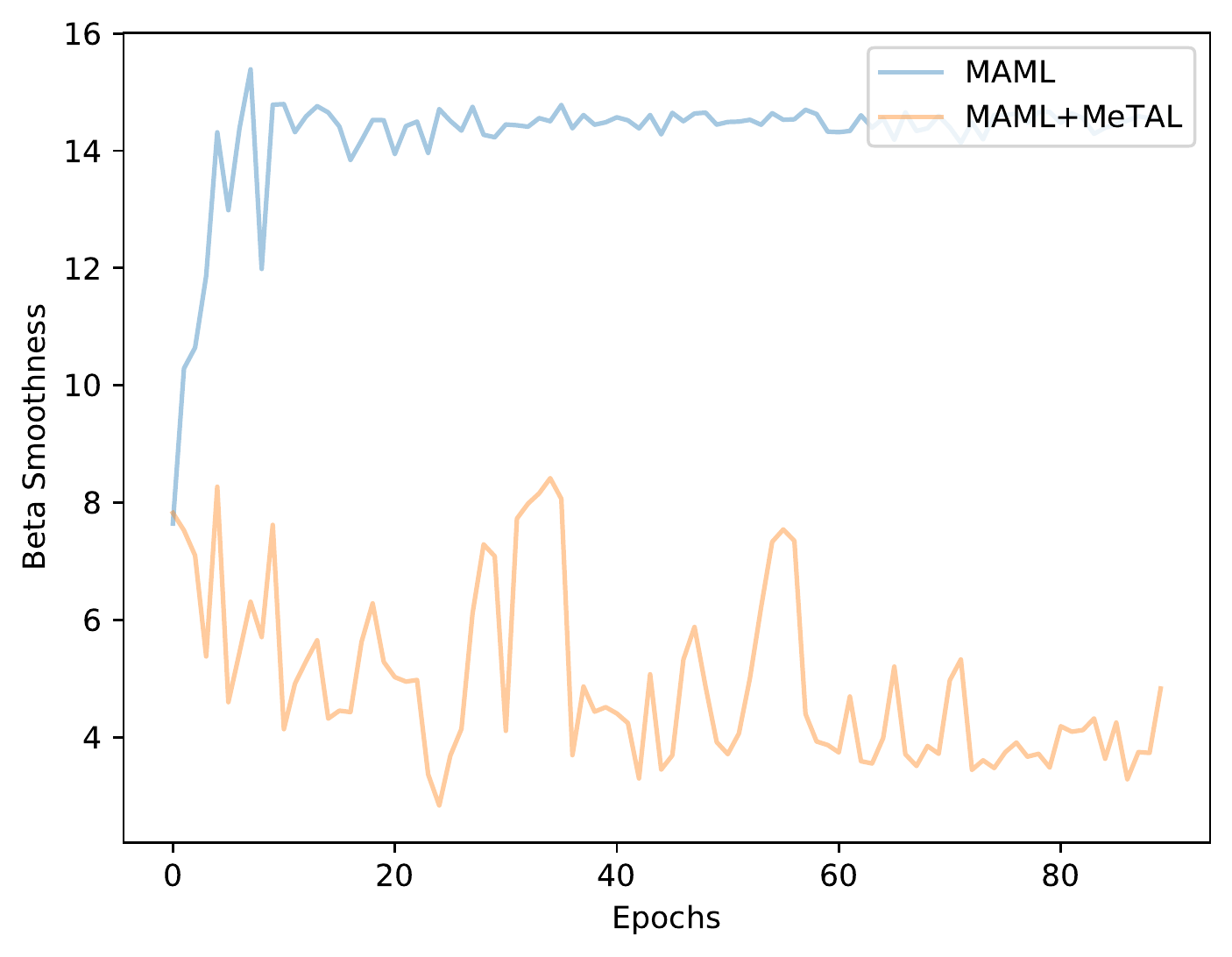}
    \label{fig:beta}
}
\end{center}
\vspace{-2em}
\caption{{Inner-loop loss landscape visualization}}
\label{fig:landscape}
\end{figure*}

To further stress the effectiveness of MeTAL, we analyze and compare the loss landscapes from MAML~\cite{finn2017model} and MeTAL.
To this end, we employ a loss landscape visualization scheme~\cite{santurkar2018how} used for MAML analysis in~\cite{baik2020learning}.
Santurkar\etal~\cite{santurkar2018how} analyze the stability and smoothness of optimization landscape by measuring loss variations (\ie loss landscape), changes in gradients (\ie gradient predictiveness), and the maximum difference in gradients (\ie ``effective'' $\beta$-smoothness).
Figure~\ref{fig:landscape} demonstrates loss variations (a), changes in gradients (b), and the maximum difference in gradients (c) are measured for meta-validation set at the first inner-loop step and averaged for each meta-training epoch on 5-way 5-shot miniImageNet classification.
The thinner shades in (a), (b), and the lower value in (c) indicate the smoother loss landscape.
MeTAL (\textcolor{YellowOrange}{orange}) shows a relatively smoother loss landscape (\ie the learned loss function is relatively well-behaved), compared to MAML (\textcolor{NavyBlue}{blue}). 

\section{Few-Shot Classification}
In this section, we provide more few-shot classification results on CIFAR100-derived~\cite{krizhevsky2009learning} benchmarks and more thorough comparisons with the recent optimization-based meta-learning algorithms.

\subsection{CIFAR100-based Datasets}
In addition to Table \textcolor{red}{1} in the main text, we further validate the effectiveness of MeTAL on other few-shot classification benchmarks: namely, CIFAR-FS~\cite{bertinetto2019meta} and FC100~\cite{oreshkin2018tadam}.
Both CIFAR-FS and FC100 are derived from CIFAR100~\cite{krizhevsky2009learning} and composed of images with low-resolution of $32\times32$.
Bertinetto~\etal~\cite{bertinetto2019meta} uses a procedure similar to miniImageNet~\cite{ravi2017optimization} and randomly samples and splits the original CIFAR100 dataset to obtain CIFAR-FS.
On the other hand, FC100~\cite{oreshkin2018tadam} is obtained by using a dataset construction process similar to tieredImageNet~\cite{ren2018meta}, in which class hierarchies are used to split the original dataset to simulate more challenging few-shot learning scenarios.
The overall results and comparisons with recent optimization-based meta-learning algorithms on the two datasets are presented in Table~\ref{tab:result_cifar}.

The table illustrates that, for the same base learner backbone (4-CONV or ResNet12), MeTAL proves a state-of-the-art performance when used together with ALFA~\cite{baik2020meta}, which is a recently introduced inner-loop optimizer for optimization-based meta-learning algorithms.
Specifically, MeTAL is shown to outperform even recent methods that use pretrained networks or larger networks (WRN-28-10), especially on 5-shot classification.
Note that SIB~\cite{hu2020empirical} and SIB + E3BM~\cite{liu2020ensemble} also attempt to explicitly utilize transductive setting, similar to MeTAL.
MeTAL (with the same backbone or sometimes even smaller backbone) outperforms SIB and its variants, which use pretrained networks, suggesting that the task-adaptive loss function by MeTAL effectively extracts useful information from the unlabeled examples.
Furthermore, MeTAL exhibits similar tendency that can be observed in Table\textcolor{red}{1} in the main text: MeTAL provides consistent performance improvement across different baselines, base learner backbone, and datasets.
These experimental results reinforce our claim that learning a good loss function, which has been significantly less explored, is just as beneficial for generalization as learning a good initialization or a good optimizer.

\label{cifar}
\begin{table*}[h!]
\begin{center}
   \centering 
   \scalebox{0.95}{
   \begin{threeparttable}
   \begin{tabular}{lcc@{\hspace{-0.2cm}}ccc@{\hspace{-0.2cm}}ccc}
   \toprule[\heavyrulewidth]
          \multirow{2}{*}{\textbf{Model}}& \multirow{1}{*}{\textbf{Base learner}} & \phantom{abc}  &\multicolumn{2}{c}{\textbf{CIFAR-FS}} & \phantom{abc}& \multicolumn{2}{c}{\textbf{FC100}} \\
           \cmidrule{4-5}\cmidrule{7-8}
          &\textbf{Backbone}& &1-shot & 5-shot && 1-shot & 5-shot\\
   \midrule
   BOIL  ~\cite{oh2021boil} & 4-CONV  && $58.03 \pm 0.43\%$ & $73.61 \pm 0.32\%$ && $38.93 \pm 0.45\%$ & $51.66 \pm 0.32\%$ \\
   MAML + gcp-sampling ~\cite{liu2020adaptive} & 4-CONV  && $57.62 \pm 0.97\%$ & $72.51 \pm 0.72\%$ & &  - & - \\
   MAML++ + gcp-sampling ~\cite{liu2020adaptive} & 4-CONV  && $60.14 \pm 0.97\%$ & $73.98 \pm 0.74\%$ & &  - & - \\
   SIB * ~\cite{hu2020empirical} & 4-CONV && $68.7 \pm 0.6\%$ & $77.1 \pm 0.4\%$ & &  - & - \\
   META-RHKS-I ~\cite{zhou2021meta} &4-CONV && - & -  && $38.90 \pm 1.90\%$ & $51.47 \pm 0.86\%$ \\
   META-RHKS-II ~\cite{zhou2021meta} &4-CONV &&  - & -  && $\textbf{41.20} \pm \textbf{2.17}\%$ & $51.36 \pm 0.96\%$ \\
   MAML + ${\mathbf E}^3$BM ~\cite{liu2020ensemble} & 4-CONV &&  - & -  && $39.9 \pm 1.8\%$ & $52.6 \pm 0.9\%$ \\\hdashline \\[-8pt]
   MAML\tnote{\textdaggerdbl} & 4-CONV && $57.63 \pm 0.73\%$ & $73.95 \pm 0.84\%$ && $35.89 \pm 0.72\%$ & $49.31 \pm 0.47\%$  \\
   \textbf{MeTAL (Ours)} & 4-CONV & & $59.16 \pm 0.56\%$ & $74.62 \pm 0.42\%$ && $37.46 \pm 0.39\%$ & $51.34 \pm 0.25\%$  \\ \hdashline \\[-8pt]
   ALFA + MAML ~\cite{baik2020meta} & 4-CONV & & $59.96 \pm 0.49\%$ &  $76.79 \pm 0.42\%$ && $37.99 \pm 0.48\%$ & $53.01 \pm 0.49\%$ \\
   ALFA + \textbf{MeTAL (Ours)} & 4-CONV & & $\textbf{69.19} \pm \textbf{0.27}\%$ & $\textbf{79.33} \pm \textbf{0.28}\%$ && $\textbf{42.24} \pm \textbf{0.47}\%$ & $\textbf{55.36} \pm \textbf{0.16}\%$ \\
   \midrule
   MetaOpt \tnote{\textdagger} ~\cite{lee2019meta} & ResNet12 & & $72.0 \pm 0.7\%$ & $84.2 \pm 0.5\%$& & $41.1 \pm 0.6\% $ & $55.5 \pm 0.6\%$ \\ \hdashline \\[-8pt]
   MAML\tnote{\textdaggerdbl} & ResNet12 & & $63.81 \pm 0.54\%$ & $77.07 \pm 0.42\%$ && $37.29 \pm 0.40\%$ & $50.70 \pm 0.35\%$\\
   \textbf{MeTAL (Ours)} & ResNet12 & & $67.97 \pm 0.47\%$ & $82.17 \pm 0.38\%$ && $39.98 \pm 0.39\%$ & $53.85 \pm 0.36\%$ \\ \hdashline \\[-8pt]
   ALFA + MAML ~\cite{baik2020meta} & ResNet12 & & $66.79 \pm 0.47\%$ & $83.62 \pm 0.37\%$ && $41.46 \pm 0.49\%$ & $55.82 \pm 0.50\%$ \\
   ALFA + \textbf{MeTAL (Ours)} & ResNet12 & &  $\textbf{76.32} \pm \textbf{0.43}\%$ & $\textbf{86.73} \pm \textbf{0.31}\%$ && $\textbf{44.54} \pm \textbf{0.50}\%$ & $\textbf{58.44} \pm \textbf{0.42}\%$ \\
   \midrule
   SIB * ~\cite{hu2020empirical} & WRN-28-10 && $80.0 \pm 0.6\%$ & $85.3 \pm 0.4\%$ & &  - & - \\
   SIB + ${\mathbf E}^3$BM$^*$ ~\cite{liu2020ensemble} & WRN-28-10 &&  - & -  && $46.0 \pm 0.6\%$ & $57.1 \pm 0.4\%$ \\
   \bottomrule[\heavyrulewidth] 
   \end{tabular}
   \begin{tablenotes}
   \small\item[*] Pretrained
   \small\item[\textdagger] Trained with data augmentation.
   \small\item[\textdaggerdbl] Reproduced.
   \end{tablenotes}
   \end{threeparttable}
   }
   \caption{5-way 1-shot and 5-way 5-shot classification test accuracy on CIFAR-based datasets: CIFAR-FS and FC100.}
   \label{tab:result_cifar}
\end{center}
\end{table*}

\subsection{Detailed comparisons on ImageNet-based Datasets}
We augment Table \textcolor{red}{1} in the main text with more comparisons with other recent optimization-based meta-learners, as presented in Table~\ref{tab:result_imagenet_supp}.
Similar to results on CIFAR-based benchmarks, MeTAL is shown to outperform most recent methods with the similar base learner backbone (4-CONV or ResNet12), including methods that utilize pretrained feature extractor, which may limit their application or effectiveness to classification problems only.
Providing the competitive performance without relying on pretraining or data augmentation, our proposed method MeTAL demonstrates its effectiveness in achieving generalization.

\begin{table*}[h!]
\begin{center}
   \centering 
   \scalebox{0.95}{
   \begin{threeparttable}
   \begin{tabular}{lcc@{\hspace{-0.2cm}}ccc@{\hspace{-0.2cm}}ccc}
   \toprule[\heavyrulewidth]
          \multirow{2}{*}{\textbf{Model}}& \multirow{1}{*}{\textbf{Base learner}} & \phantom{abc}  &\multicolumn{2}{c}{\textbf{miniImageNet}} & \phantom{abc}& \multicolumn{2}{c}{\textbf{tiredImageNet}} \\
           \cmidrule{4-5}\cmidrule{7-8}
          &\textbf{Backbone}& &1-shot & 5-shot && 1-shot & 5-shot\\
   \midrule
   MAML + ${\mathbf E}^3$BM ~\cite{liu2020ensemble} & 4-CONV && $53.2 \pm 1.8\%$ & $65.1 \pm 0.9\%$ && $52.1 \pm 1.8\%$ & $70.2 \pm 0.9\%$ \\
   Meta-RHKS-I ~\cite{zhou2021meta} & 4-CONV  && $51.10 \pm 1.82\%$ & $66.19 \pm 0.80\%$ && - & - \\
   Meta-RHKS-II ~\cite{zhou2021meta} & 4-CONV  && $50.03 \pm 2.09\%$ & $65.40 \pm 0.91\%$ && - & - \\
   BOIL ~\cite{oh2021boil} & 4-CONV  && $49.61 \pm 0.16\%$ & $66.45 \pm 0.37\%$ && $48.58 \pm 0.27\%$ & $69.37 \pm 0.12\%$ \\
   MAML + Meta-Dropout ~\cite{lee2020meta} & 4-CONV  && $51.93 \pm 0.67\%$ & $67.42 \pm 0.52\%$ && - & - \\
   Meta-SGD + Meta-Dropout ~\cite{lee2020meta} & 4-CONV  && $50.87 \pm 0.63\%$ & $65.55 \pm 0.57\%$ && - & - \\
   ModGrad ~\cite{grant2020modulating} & 4-CONV  && $53.20 \pm 0.86\%$ & $69.17 \pm 0.69\%$ & & \\
   MAML + gcp-sampling ~\cite{liu2020adaptive} & 4-CONV  && $49.65 \pm 0.85\%$ & $65.37 \pm 0.70\%$ && - & - \\
   MAML++ + gcp-sampling ~\cite{liu2020adaptive} & 4-CONV  && $52.34 \pm 0.81\%$ & $69.21 \pm 0.68\%$ && - & - \\
   MAML + L2F ~\cite{baik2020learning} & 4-CONV  && $52.10 \pm 0.50\%$ & $69.38 \pm 0.46\%$ && $54.40 \pm 0.50\%$ & $73.34 \pm 0.44\%$ \\
   
   SIB$^*$ ~\cite{hu2020empirical} & 4-CONV  && $\textbf{58.0} \pm \textbf{0.6}\%$ & $70.7 \pm 0.4\%$ && - & - \\ \hdashline \\[-8pt]
   MAML\tnote{\textdaggerdbl} & 4-CONV && $49.64\pm 0.31\%$ & $64.99 \pm 0.27\%$ &&$50.98\pm 0.26\%$ & $66.25 \pm 0.19\%$ \\
   \textbf{MeTAL (Ours)} & 4-CONV & & $52.63 \pm 0.37\%$ & $70.52 \pm 0.29\%$ &&$54.34\pm 0.31\%$ & $70.40 \pm 0.21\%$ \\ \hdashline \\[-8pt]
   ALFA + MAML ~\cite{baik2020meta} & 4-CONV & & $50.58 \pm 0.51\%$ & $69.12 \pm 0.47\%$ & &$53.16 \pm 0.49\%$ & $70.54 \pm 0.46\%$ \\
   ALFA + \textbf{MeTAL (Ours)} & 4-CONV & & $\textbf{57.75} \pm \textbf{0.38}\%$ & $\textbf{74.10} \pm \textbf{0.43}\%$ & &$\textbf{60.29} \pm \textbf{0.37}\%$ & $\textbf{75.88} \pm \textbf{0.29}\%$  \\
   \midrule
   Warp-MAML ~\cite{flennerhag2020meta} & 4-CONV(128)\tnote{\S}  && $52.3 \pm 0.8\%$ & $68.4 \pm 0.6\%$ && $57.2 \pm 0.9\%$ & $74.1 \pm 0.7\%$ \\
   SIB$^*$ ~\cite{hu2020empirical} & 4-CONV(128)\tnote{\S}  && $63.26 \pm 1.07\%$ & $75.73 \pm 0.71\%$ && - & - \\
   \midrule
   MAML + L2F & ResNet12 && $57.48 \pm 0.49\%$ & $74.68 \pm 0.43\%$ && $63.94 \pm 0.48\%$ & $77.61 \pm 0.41\%$ \\
   MetaOpt \tnote{\textdagger} ~\cite{lee2019meta} & ResNet12 & &  $62.64 \pm 0.61\% $ & $78.63 \pm 0.46 \%$ && $65.99 \pm 0.72\%$ & $81.56 \pm 0.53\%$ \\
   SIB + IFSL$^*$ \tnote{\textdagger} ~\cite{yue2020interventional} & ResNet10  && $\textbf{67.10} \pm \textbf{0.56}\%$ & $78.88 \pm 0.35\%$ && $\textbf{77.64} \pm \textbf{0.58}\%$ & $85.09 \pm 0.35\%$ \\\hdashline \\[-8pt]
   MAML\tnote{\textdaggerdbl} & ResNet12 & & $58.60 \pm 0.42\%$ & $69.54 \pm 0.38\%$ & &$59.82 \pm 0.41\%$ & $73.17 \pm 0.32\%$\\
   \textbf{MeTAL (Ours)} & ResNet12 & & $59.64 \pm 0.38\%$ & $76.20 \pm 0.19\%$ & &$63.89\pm 0.43\%$ & $80.14 \pm 0.40\%$\\ \hdashline \\[-8pt]
   ALFA + MAML ~\cite{baik2020meta} & ResNet12 & &  $59.74 \pm 0.49\%$ & $77.96 \pm 0.41\%$&& $64.62 \pm 0.49\%$ & $82.48 \pm 0.38\%$ \\
   ALFA + \textbf{MeTAL (Ours)} & ResNet12 & &   $66.61 \pm 0.28\%$ & $\textbf{81.43} \pm \textbf{0.25}\%$ & & $70.29 \pm 0.40\%$ &$\textbf{86.17} \pm \textbf{0.35}\%$\\
   \midrule
   LEO-trainval$^*$~\cite{rusu2019meta} & WRN-28-10 & & $61.76 \pm 0.08\% $ & $77.59 \pm 0.12 \%$ && $66.33 \pm 0.05\%$ & $81.44 \pm 0.09\%$ \\
   LEO + L2F$^*$ ~\cite{baik2020learning} & WRN-28-10  && $62.12 \pm 0.13\%$ & $78.13 \pm 0.15\%$ && $68.00 \pm 0.11\%$ & $83.02 \pm 0.08\%$ \\
   SIB$^*$ ~\cite{hu2020empirical} & WRN-28-10  && $70.0 \pm 0.6\%$ & $79.2 \pm 0.4\%$ && - & - \\
   ModGrad ~\cite{grant2020modulating} & WRN-28-10 &&  $65.72 \pm 0.21\%$ & $81.17 \pm 0.20\%$ && - & - \\
   SIB + ${\mathbf E}^3$BM$^*$ ~\cite{liu2020ensemble} & WRN-28-10  && $71.4 \pm 0.5\%$ & $81.2 \pm 0.4\%$ && $75.6 \pm 0.6\%$ & $84.3 \pm 0.4\%$ \\
   SIB + IFSL$^*$ \tnote{\textdagger} ~\cite{yue2020interventional} & WRN-28-10 && $\textbf{71.31} \pm \textbf{0.56}\%$ & $\textbf{81.73} \pm \textbf{0.34}\%$ && $\textbf{81.97} \pm \textbf{0.56}\%$ & $\textbf{88.19} \pm \textbf{0.34}\%$ \\
   \bottomrule[\heavyrulewidth] 
   \end{tabular}
   \begin{tablenotes}
   \small\item[*] Pretrained
   \small\item[\textdagger] Trained with data augmentation.
   \small\item[\textdaggerdbl] Reproduced.
   \small\item[\S] Larger 4-CONV architecture with 128 filters.
   \end{tablenotes}
   \end{threeparttable}
   }
      \caption{5-way 1-shot and 5-way 5-shot classification test accuracy on ImageNet-based datasets: miniImageNet and tieredImageNet.}
   \label{tab:result_imagenet_supp}
\end{center}
\end{table*}

\section{Semi-Supervised Inner-Loop Optimization}
\begin{table*}[h!]
   \centering 
    \begin{tabular}{ccccc}
        \toprule
          \verb|#| query & \verb|#| non-query & \verb|#| distractor& MeTAL & ALFA+MeTAL \\\midrule
         15  &    0    &     0   &  $70.52 \pm 0.29\%$  & $74.10 \pm 0.43\%$\\\hdashline \\[-8pt]
         0 & 5 &   0 &  $68.90 \pm  0.39\%$ & $71.62 \pm  0.34\%$\\
         0 & 10 &   0 &  $69.76 \pm  0.48\%$ & $72.33 \pm  0.27\%$\\
         0 & 15 &   0 &  $70.40 \pm  0.34\%$ & $73.48 \pm  0.20\%$\\
         5 & 10 &   0 &  $70.02 \pm  0.45\%$ & $72.98 \pm  0.32\%$\\
         10 & 5 &   0 &  $70.06 \pm  0.41\%$ & $73.21 \pm  0.36\%$\\ \hdashline \\[-8pt]
         0 & 0 &   5 &   $67.69 \pm  0.39\%$ & $69.72 \pm  0.45\%$\\
         0 & 0 &   10 &  $67.02 \pm  0.48\%$ & $70.02 \pm  0.49\%$\\
         0 & 0 &   15 &  $66.97 \pm  0.34\%$ & $70.63 \pm  0.46\%$\\
         5 & 5 &   5 &   $67.58 \pm  0.47\%$ & $71.50 \pm  0.45\%$\\
        \bottomrule
    \end{tabular}
    \caption{Investigation on the effectiveness of MeTAL under various semi-supervised few-shot classification scenarios. In addition to support examples, different combination of unlabeled examples are used in inner-loop optimization. miniImageNet 5-\textit{way} 5-\textit{shot} classification accuracy is reported with a 4-CONV backbone.
    \# implies the number of unlabeled examples per each \textit{way}.
    Each distractor image is sampled from a different class.
    }
    \label{tab:semi-supervised}
\end{table*}

Recently, metric-based meta-learning algorithms, such as the method from~\cite{liu2018learning}, have attempted to make full use of the \textit{unlabeled query} set by exploiting its feature similarity with the labeled support set.
Under such scenario (\textit{a.k.a.} \textit{transductive} setting), many recent metric-based meta-learning algorithms have achieved outstanding performance.
On the other hand, the transductive setting or transductive inference is rarely explored among optimization-based learners.
Recent few works~\cite{antoniou2019learning,hu2020empirical} have applied transductive inference to optimization-based methods to utilize the information available from the \textit{unlabeled query} set.
However, these works only explore finetuning to the given query set, without considering a scenario, where they may exist a batch of unlabeled data prior to the inference or meta-test time~\cite{ren2018meta}.
We perform a small ablation study that shows MeTAL does not learn to finetune to the given query set but rather learns to extract information from the unlabeled images. 

To this end, we introduce a semi-supervised few-shot classification setting, similar to~\cite{ren2018meta}.
Similar to how Ren~\etal~\cite{ren2018meta} has set up semi-supervised few-shot classification, we divide unlabeled data into query set, non-query set, and distractor set.
Query set is a set of examples whose classes are to be estimated.
Non-query set is a set of examples that belong to the same task (same set of classes) as the query set.
The difference from the query set is that the non-query set is available before the inference time or query set is given 
Distractor set is a set of examples that belong to different tasks (different classes).

Table~\ref{tab:semi-supervised} reports the 5-way 5-shot classification test accuracy with a 4-CONV base learner backbone on miniImageNet when various combinations of three types of unlabeled examples are used, instead of original 15 query examples per class (first row in the table), during the inner-loop optimization.
The table shows that the classification accuracy increases with the number of non-query unlabeled examples, implying that MeTAL learns to generalize better by extracting relevant information from the unlabeled examples, instead of finetuning/overfitting to the given set of unlabeled examples.
Surprisingly, MeTAL manages, to some extent, the performance under the presence of irrelevant or maybe even destructive distractor sets. 
In particular, the accuracy of MeTAL does not drop significantly with increasing number of distractor sets.
This further corroborates that MeTAL, unlike other methods, does not finetune or overfit to the given unlabeled examples but rather attempts to obtain better generalization.

\section{Visual Tracking}
\begin{figure}[t]
	\centering
	\includegraphics[width=\linewidth]{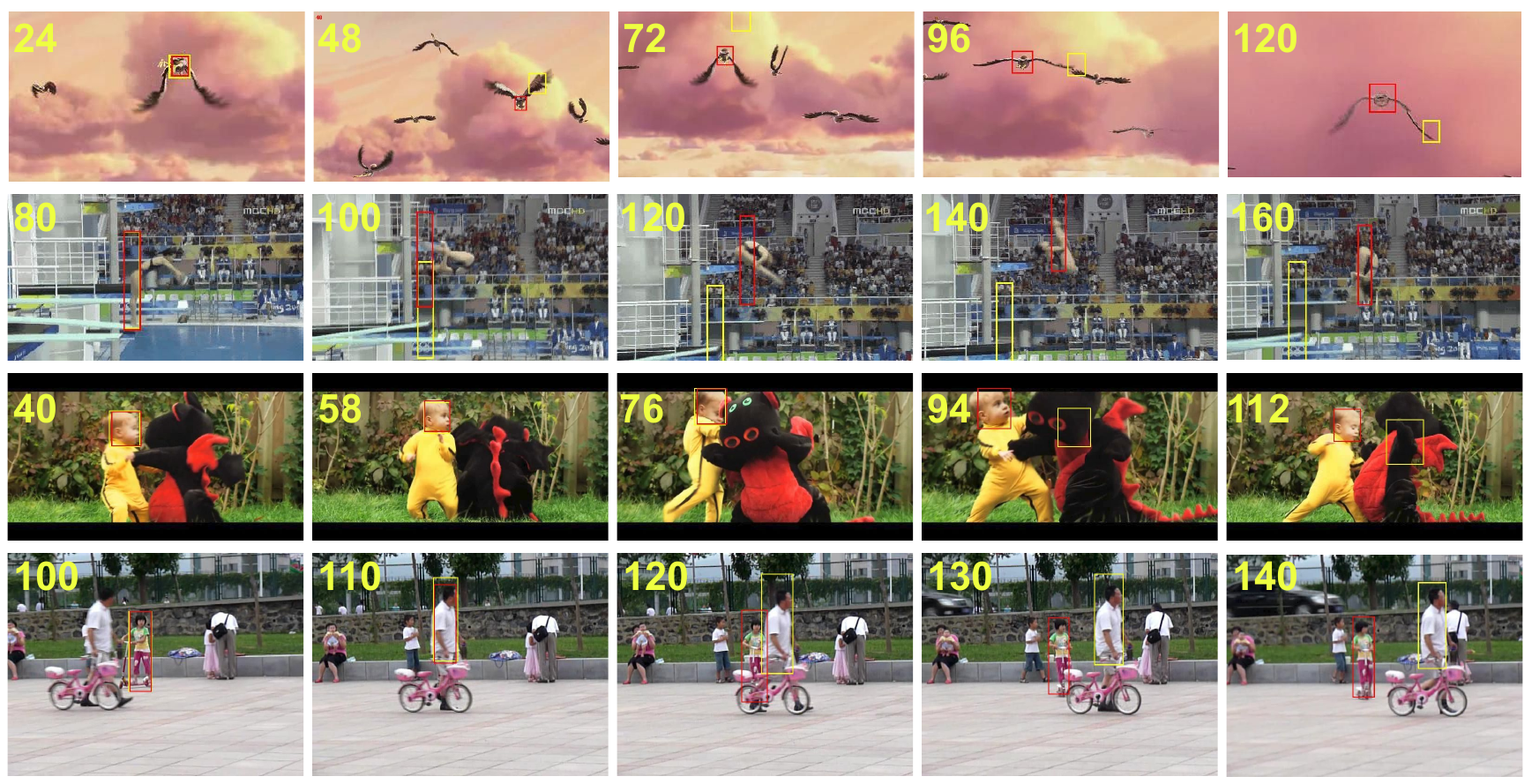}
	\caption{Examples of meta-tracking results. \textbf{\textcolor{yellow}{Yellow box}} denotes MetaCREST and \textbf{\textcolor{red}{red box}} denotes MetaCREST+MeTAL.
	Results are shown for sequences in the OTB2015 dataset where each row shows selected frames from \textit{bird1}, \textit{diving}, \textit{dragonBaby}, and \textit{girl2} sequences.}
	\label{fig:meta-tracking}
\end{figure}

To further demonstrate the applicability and flexibility of MeTAL, we apply our proposed method in visual tracking.
Visual tracking is a challenging problem, in which the goal is to track the target whose bounding box is given only in the first frame of the video.
As such problem setting is inherently a few-shot learning problem, one of the most flexible few-shot learning methodologies MAML~\cite{finn2017model} has gained attention from visual tracking community.
In particular, Park~\etal~\cite{park2018meta} has employed MAML to one of the existing tracking algorithms, such as CREST~\cite{song2017crest} to better adapt to object appearance changes throughout video frames, naming the newly obtained tracker MetaCREST.
We use their publicly released code and apply MeTAL to MetaCREST to evaluate the capability and flexibility of the proposed task-adaptive loss function under more realistic and challenging scenarios, as shown in Table~\ref{tab:tracking} and Figure~\ref{fig:meta-tracking}.
Both quantitatively and qualitatively, MeTAL demonstrates performance improvement over MetaCREST, validating the effectiveness and flexibility of MeTAL in learning a loss function that provides better generalization for each task.
Note that in visual tracking, it is difficult to handle query examples (a new frame) as we could in few-shot classification or simple few-shot regression.
As such, we do not employ semi-supervised inner-loop optimization and thus only evaluate the effectiveness of a task-adaptive loss function.

\begin{table}
   \small
   \centering
   \begin{tabular}{lcc}
   \toprule[\heavyrulewidth]
   Model       &  Precision & Success rate\\
   \midrule
   MetaCREST~\cite{park2018meta} & 0.7994 & 0.6029\\
   MetaCREST + MeTAL & \textbf{0.8253} & \textbf{0.6143} \\
   \bottomrule[\heavyrulewidth] 
   \end{tabular}
    \caption{Precision and success rate measured over 100 sequences in the OTB2015 dataset~\cite{wu2015otb} by using one-pass evaluation (OPE) protocol.} 
   \label{tab:tracking}
\end{table}

\section{Visualization of Affine Transformation Parameters}

\begin{figure*}[h]
	\centering
	{\small \null \hspace{0.05\linewidth} (a) $\gamma$ for Layer 1 weight \hspace{0.05\linewidth} (b)  $\beta$ for Layer 1 weight \hspace{0.07\linewidth} (c) $\gamma$ for Layer 1 bias \hspace{0.08\linewidth} (d) $\beta$ for Layer 1 bias  \hspace{\fill}  \null}\\
	\includegraphics[width=0.24\linewidth]{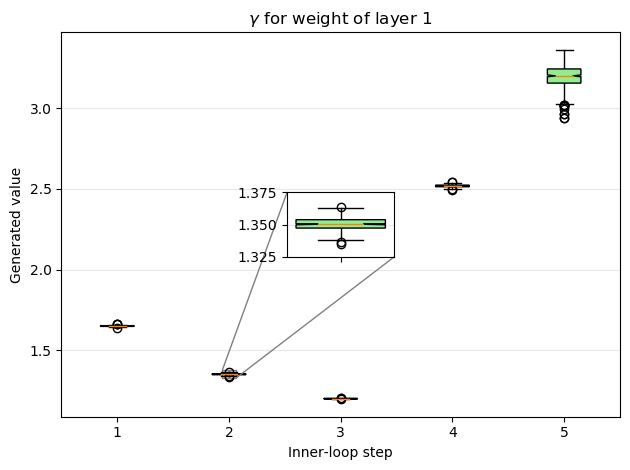}
	\includegraphics[width=0.24\linewidth]{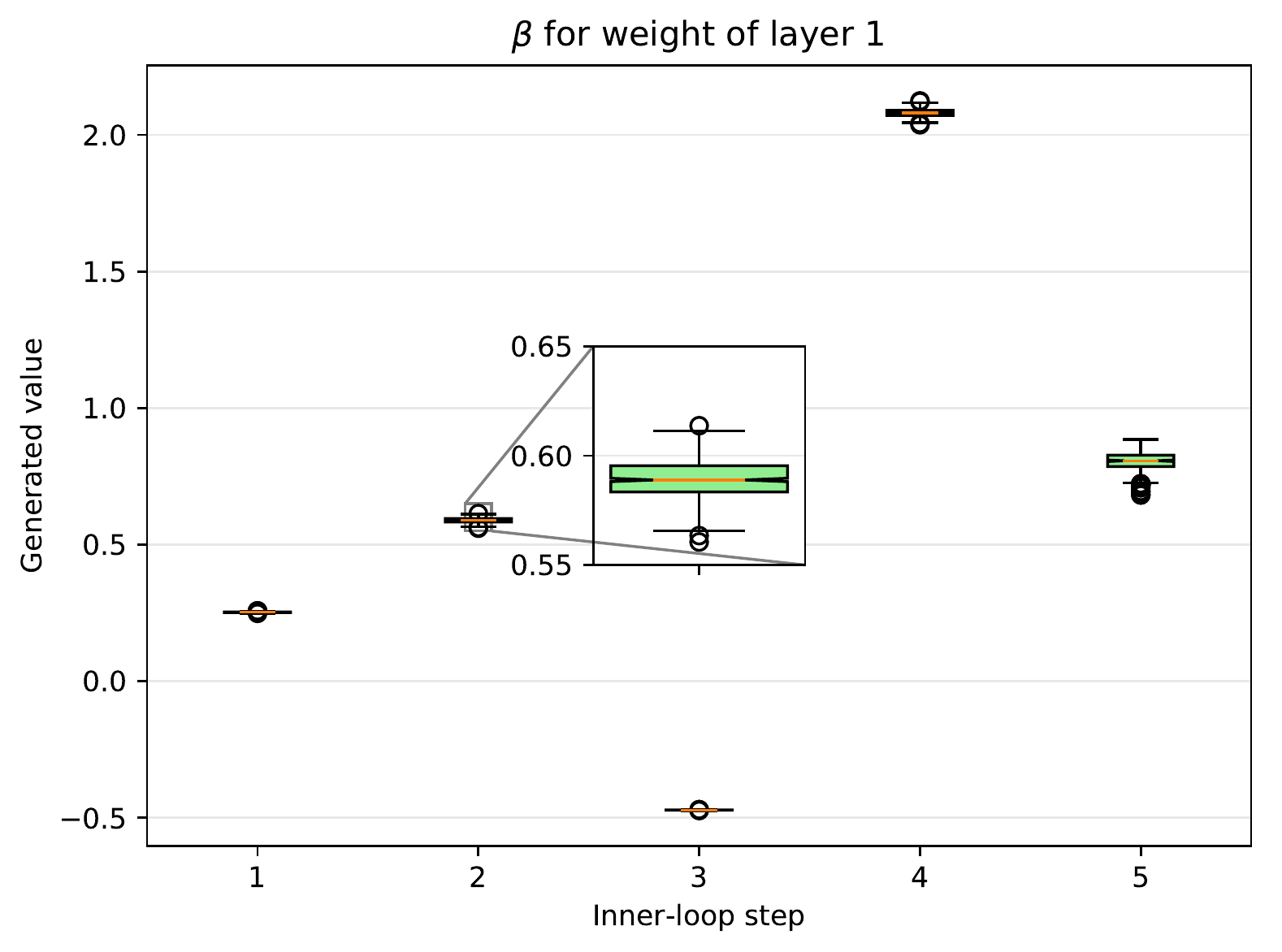}
	\includegraphics[width=0.24\linewidth]{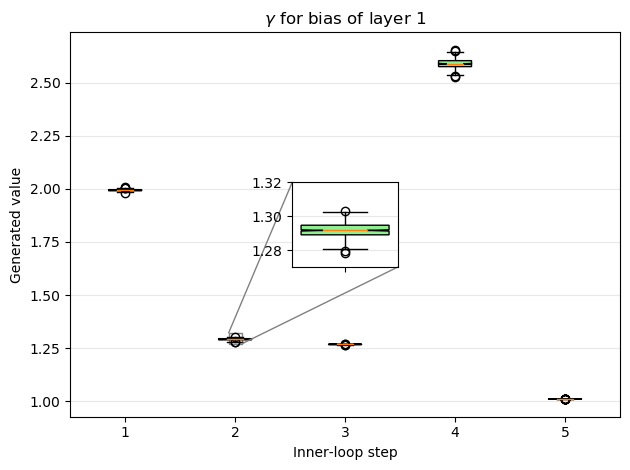}
	\includegraphics[width=0.24\linewidth]{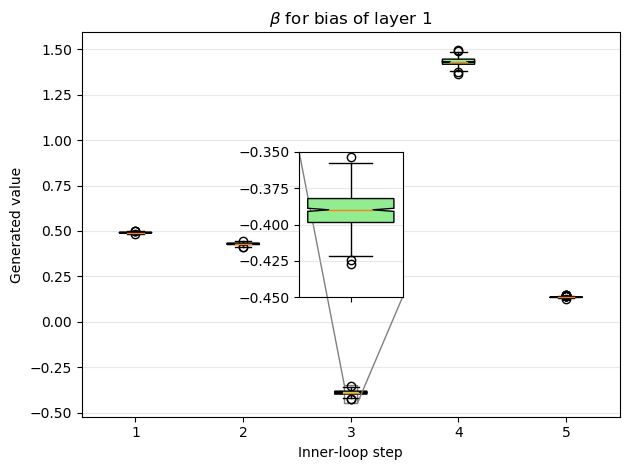}
	\\
	\centering
	{\small \null \hspace{0.05\linewidth} (a) $\gamma$ for Layer 2 weight \hspace{0.05\linewidth} (b)  $\beta$ for Layer 2 weight \hspace{0.07\linewidth} (c) $\gamma$ for Layer 2 bias \hspace{0.08\linewidth} (d) $\beta$ for Layer 2 bias  \hspace{\fill}  \null}\\
	\includegraphics[width=0.24\linewidth]{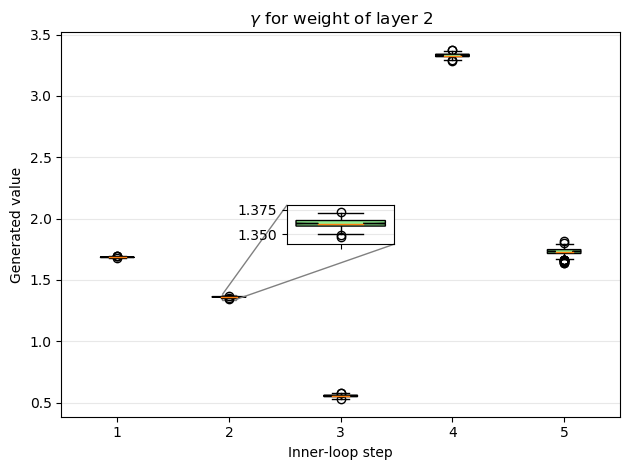}
	\includegraphics[width=0.24\linewidth]{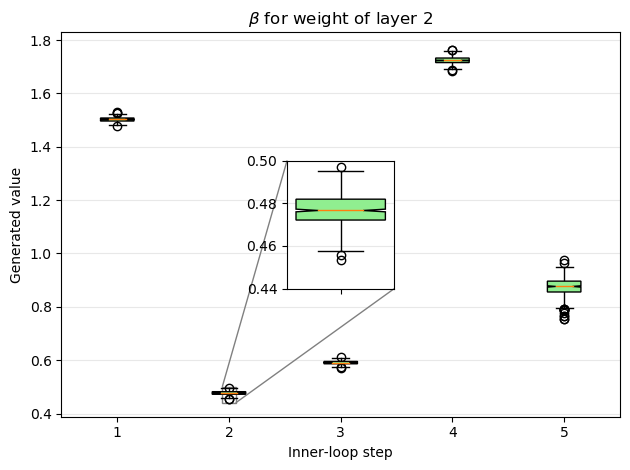}
	\includegraphics[width=0.24\linewidth]{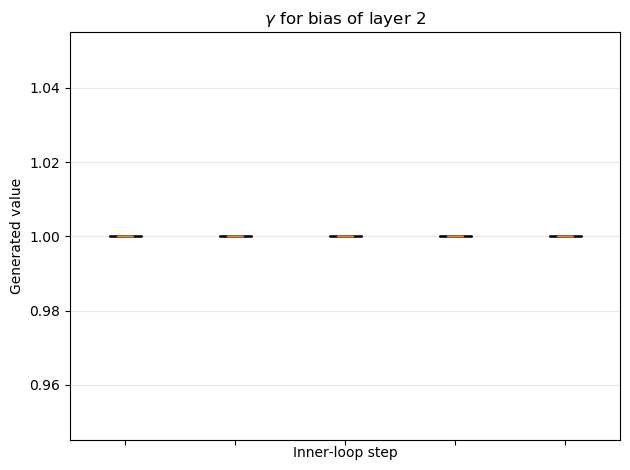}
	\includegraphics[width=0.24\linewidth]{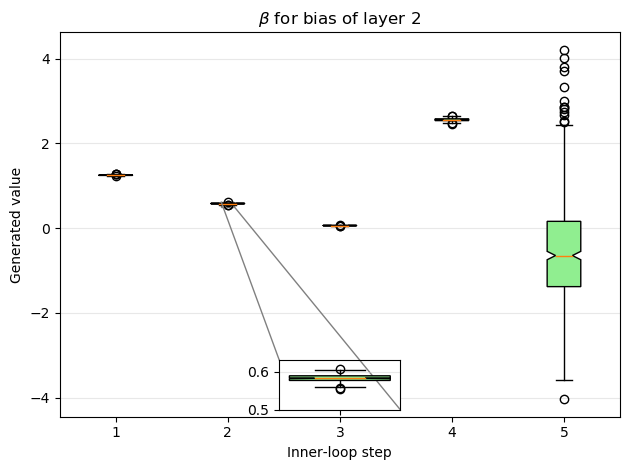}
	\caption{Visualization of affine transformation parameters generated by the meta-network $g_\psi$ across different layers of loss learner network $\mathcal{L}_\phi$, different inner-loop steps, and different validation tasks. 
	Visualization is performed on 5-way 5-shot miniImageNet validation set.} 
	\label{fig:supp_viz}
	\vspace{-0.2cm}
\end{figure*}

In addition to Figure \textcolor{red}{2} in the main text, we illustrate the affine transformation parameters generated by our proposed adapter meta-network $g_\psi$ for other loss learner network parameters in Figure~\ref{fig:supp_viz}.
Exhibit consistent tendency with Figure \textcolor{red}{2}, MeTAL demonstrates dynamic behaviour across inner-loop steps and tasks.
Interestingly, MeTAL learns to dynamically change the offset ($\beta$) of the second layer bias while minimizing the scaling.

\section{Implementation Details}
For experiments on $N$-way $k$-shot classification in this work, we follow the typical settings that are similar to~\cite{finn2017model} when training and reporting results for our baselines (MAML~\cite{finn2017model}, MAML++~\cite{antoniou2019how}, ALFA~\cite{baik2020meta}) and our method MeTAL. 
During both meta-training and evaluation, inner-loop optimization (\textit{a.k.a.} fast adaptation) is performed with a fixed number (5 in this work) of inner-loop steps with an inner-loop learning rate of  $\alpha = 0.1$.
Meta-training for each reproduced baseline and our method is performed with second-order gradients and a meta-learning rate of $\eta = 0.001$ for 100 epochs, each of which has 500 iterations.
As with the typical settings~\cite{finn2017model,antoniou2019how,baik2020meta}, each task consists of $15$ query examples ($15$ \textit{shots}) per class (hence 75 in total for 5-way classification: $|\mathcal{D}^Q_i|$ = 75) for both meta-training and evaluation.
Again, as with previous works, models are trained with a meta-batch size of $2$ for $5$-shot and $4$ for $1$-shot classification.
Similar to~\cite{antoniou2019how,baik2020meta}, all results reported in this work are obtained by an ensemble of $5$ top-validation-performance models from the same run, the whole process of which is repeated 3 times with different random seeds.

For a base learner network $f$, we adopt an architecture design from~\cite{antoniou2019how,ravi2017optimization,finn2017model,baik2020meta} for 4-CONV and ~\cite{oreshkin2018tadam,baik2020learning,baik2020meta} for ResNet12.
In particular, 4-CONV has $4$ convolution layers with a fully-connected layer and softmax at the end for classification.
Each convolution layer is composed of $48$ convolution filters of size $3\times3$, a batch normalization~\cite{ioffe2015batch} unit, a Leaky ReLU non-linear activation unit, and a max pooling layer of size $2\times2$. 
ResNet12 has $4$ residual blocks with, again, a fully-connected layer and softmax at the end for classification.
Each residual block is composed of three convolution operations, each with a filter size of $3 \times 3$.
In between convolution operations, a batch normalization unit and a ReLU non-linear activation unit are placed.
At the end of each residual block has a sequence of a batch normalization unit, a skip connection, a ReLU non-linear activation unit, and a max pooling unit of size $2\times2$.
A skip connection itself has a batch normalization unit and a ReLU non-linear activation unit.
The first residual block has $64$ filters for each convolution operation, with each successive residual block having double the number of filters from a preceding block.
Each experiment with 4-CONV base learner backbone is performed on a single NVIDIA GeForce GTX 2080Ti GPU while ResNet12 base learner backbone on a single NVIDIA Quadro RTX 8000 GPU.

For the proposed loss adapter meta-network $g_{\bm{\psi}}$, we employ a 2-layer MLP with ReLU activation unit between layers, as described in the main text.
The dimensions of input and hidden units are the same ($1 + L + N$), where $L$ is the number of layers of a base learner backbone network $f$ and $N$ is the dimension of the output of a base learner $f$.
The input dimension is $1 + L + N$ as the meta-network takes in a typical classical loss value $\mathcal{L}$ (cross entropy for classification), the layer-wise mean of base learner network weights, and the output of base learner $f$.
For semi-supervised settings, when feeding unlabeled query information into the meta-network, it should match the input dimension of $1 + L + N$.
Because unlabeled query examples lack ground-truth, cross entropy loss cannot be obtained.
To replace the cross entropy loss, we calculate entropy of the output of base learner to replace cross entropy loss in the case of unlabeled examples.
The output dimension of $g_{\bm{\psi}}$ is $4L_{\bm{\phi}} = 8$, generating affine transformation parameters $\bm{\gamma}$, $\bm{\beta}$ for weight and bias of each layer of loss leaner meta-network $\mathcal{L}_{\bm{\phi}}$ that has 2 layers ($L_{\bm{\phi}} = 2$).
As we desire to produce one set of affine transformation parameters for the whole task, not for each example, we take a batch-wise mean of the input such that the output has a batch dimension of $1$ (one set of affine transformation parameters).
A similar architecture design is employed for loss learner meta-network $\mathcal{L}_{\bm{\phi}}$: a 2-layer MLP with ReLU activation unit between layers.
The network input and hidden unit dimension is $1 + L + N$ while the output dimension is $1$.
As the network needs to produce $1$-dimensional scalar value for backpropagation operation (using autograd package in PyTorch library~\cite{pytorch}), we took a batch-wise mean of its output to reduce the dimension to $1$ (Note this is in contrast to the loss adapter meta-network $g_{\bm{\psi}}\psi$ that takes a batch-wise mean of the input).
For the best performance, each pair of meta-networks $g_{\bm{\psi}}\psi$ and $\mathcal{L}_{\bm{\phi}}$ has different meta-parameters for support and query examples and for each inner-loop step.
This does not increase the number of parameters significantly and boosts the performance by $1\sim2\%$.
As for regression, similar network designs are used for the two meta-networks.
For more details, please refer to the released code\footnote{The code is available at \url{https://github.com/baiksung/MeTAL}}.

\clearpage
{\small
\bibliographystyle{ieee_fullname}
\bibliography{main}
}